\documentclass{article}
\usepackage{arxiv}
\usepackage[T1]{fontenc}
\usepackage[utf8]{inputenc}
\usepackage{lmodern}
\usepackage[numbers]{natbib}
\usepackage{graphicx}
\usepackage{placeins}
\usepackage{subfig}
\usepackage{amsmath,amssymb,amsfonts}
\usepackage{mathrsfs}
\usepackage{xcolor}
\usepackage{textcomp}
\usepackage{manyfoot} 
\usepackage{bm}
\usepackage{algpseudocode}
\usepackage{listings}
\usepackage{tabularx}
\usepackage{booktabs}
\usepackage{multirow}
\usepackage{makecell}

\usepackage{siunitx}
\usepackage{colortbl}
\usepackage{caption}
\usepackage{lineno}
\usepackage{acronym}
\usepackage[utf8]{inputenc}
\usepackage[many]{tcolorbox}
\usepackage{algorithm}
\usepackage{float}

\usepackage{xurl}

\usepackage{pifont}  
\usepackage{array} 
\usepackage{tikz}
\newcommand{\mcirc}[1]{%
\tikz[baseline=(char.base)]{
  \node[draw,circle,inner sep=1pt] (char) {$#1$};}}
\usetikzlibrary{arrows.meta,positioning,calc}

\definecolor{cNavy}{HTML}{0B1F3A}
\definecolor{cBlue}{HTML}{1F6FEB}
\definecolor{cGreen}{HTML}{2EA043}
\definecolor{cRed}{HTML}{D73A49}
\definecolor{cGray}{HTML}{6B7280}
\definecolor{cBG}{HTML}{F8FAFC}

\definecolor{lightblue}{RGB}{173,216,230}
\definecolor{lightgreen}{RGB}{144,238,144}

\DeclareMathOperator*{\argmin}{\arg\!\min}
\DeclareMathOperator*{\argmax}{\arg\!\max}

\newcolumntype{L}{>{\raggedright\arraybackslash}X}
\newcolumntype{C}{>{\centering\arraybackslash}X}

\newcommand{\cmark}{\ding{51}} 
\newcommand{\xmark}{\ding{55}} 

\usepackage[
    colorlinks=true,
    linkcolor=blue,
    citecolor=blue!60!black,
    urlcolor=blue!50!black
]{hyperref}

\title{Robust multi-task boosting using clustering and local ensembling}

\author{
  Seyedsaman Emami \\
  Escuela Polit\'ecnica Superior \\
  Universidad Aut\'onoma de Madrid \\
  Madrid\\
  \And
  Daniel Hern\'andez-Lobato \\
  Escuela Polit\'ecnica Superior \\
  Universidad Aut\'onoma de Madrid \\
  Madrid\\
  Centro de Investigaci\'on Avanzada en F\'isica Fundamental \\
  Universidad Aut\'onoma de Madrid \\
  Madrid \\
   \And
  Gonzalo Mart\'{\i}nez-Mu\~noz \\
  Escuela Polit\'ecnica Superior \\
  Universidad Aut\'onoma de Madrid \\
  Madrid\\
  Centro de Investigaci\'on Avanzada en F\'isica Fundamental \\
  Universidad Aut\'onoma de Madrid \\
  Madrid \\
}

\begin{document}
\maketitle
\begin{abstract}

      Multi-Task Learning (MTL) aims to boost predictive performance by sharing
      information across related tasks, yet conventional methods often suffer from
      \emph{negative transfer} when unrelated or noisy tasks are forced to share
      representations. We propose Robust Multi-Task Boosting using Clustering and
      Local Ensembling (RMB-CLE), a principled MTL framework that integrates
      error-based task clustering with local ensembling. Unlike prior work that
      assumes fixed clusters or hand-crafted similarity metrics, RMB-CLE derives
      inter-task similarity directly from \emph{cross-task errors}, which admit a
      risk decomposition into functional mismatch and irreducible noise, providing a
      theoretically grounded mechanism to prevent negative transfer. Tasks are
      grouped adaptively via agglomerative clustering, and within each cluster, a
      local ensemble enables robust knowledge sharing while preserving task-specific
      patterns. Experiments show that RMB-CLE recovers ground-truth clusters in
      synthetic data and consistently outperforms multi-task, single-task, and
      pooling-based ensemble methods across diverse real-world and synthetic
      benchmarks. These results demonstrate that RMB-CLE is not merely a combination
      of clustering and boosting but a general and scalable framework that
      establishes a new basis for robust multi-task learning.
\end{abstract}


\section*{List of Abbreviations}
\label{sec:abbreviations}

\begin{tcolorbox}[colframe=black!30!white, colback=white, coltitle=black!70!black, boxrule=0.2mm, rounded corners, width=0.9\textwidth, grow to right by=0.1\textwidth]
      \begin{acronym}[TLC]\footnotesize
            \acro{CD}{Critical Distance}
            \acro{CMTL}{Clustered Multi-Task Learning}
            \acro{DNN}{Deep Neural Network}
            \acro{DP}{Data Pooling}
            \acro{GB}{Gradient Boosting}
            \acro{LGBM}{Light Gradient-Boosting Machine}
            \acro{MAE}{Mean Absolute Error}
            \acro{ML}{Machine Learning}
            \acro{MLP}{Multi-Layer Perceptron}
            \acro{MT}{Multi-Task}
            \acro{MTGB}{Multi-Task Gradient Boosting}
            \acro{MTL}{Multi-Task Learning}
            \acro{RMB-CLE}{Robust Multi-Task Boosting using Clustering and Local Ensembling}
            \acro{RMSE}{Root Mean Squared Error}
            \acro{R-MTGB}{Robust-Multi-Task Gradient Boosting}
            \acro{ST}{Single-Task}
            \acro{Std Dev}{Standard Deviation}
            \acro{TaF}{Task-as-Feature}
            \acro{TS}{Task-Specific}
      \end{acronym}
\end{tcolorbox}

\section{Introduction}~\label{sec:intro}
\ac{MTL} is a key domain in \ac{ML} where models are
trained to perform several related or unrelated tasks simultaneously,
facilitating the transfer of knowledge between tasks~\citep{Zhang2022}.~\ac{MTL}
aims to strengthen generalization by combining task-specific insights with
representations that are shared across multiple tasks~\citep{Caruana1997}. A
foundational work in~\citep{Caruana1997} introduced the concept of hard
parameter sharing in neural networks for \ac{MTL}, where a single model with
multiple output layers learns to solve several tasks concurrently. This setup
allows the shared layers to capture representations that are useful across
tasks, promoting inductive transfer during training.~\ac{MTL} models have numerous real-world applications;
for instance, they have been used to
simultaneously predict mood and stress levels of patients by leveraging
information from the entire population under study~\citep{Taylor2020}.

From an algorithmic perspective, \ac{MTL} approaches are often categorized into
five families: feature learning~\citep{Caruana1997,Liao2005,Silver2008,
      Gong2012,Han2014}, low-rank approaches~\citep{Ando2005,Chen2009,Agarwal2010},
task clustering~\citep{Thrun1996,bakker2003,Xue2007}, task relation
learning~\citep{Evgeniou2004,Parameswaran2010, Goernitz2011}, and decomposition
methods~\citep{Jalali2010,Han2015}. Among these, task clustering methods, also
known as \ac{CMTL}, have been widely explored.~\ac{CMTL} assumes that tasks can
be grouped into clusters based on their similarity, allowing the model to
capture shared structure within task groups while maintaining flexibility
across unrelated tasks~\citep{Zhou2016,Zhang2022}.

In this work, we concentrate on the third family, clustering-based multi-task
methods. Clustering, an unsupervised learning method, groups data based on
similarity measures~\citep{Backer1981,Rui2005,Rokach2005}. In the context of
\ac{CMTL}, clustering is used to group tasks instead of data points.
Hierarchical clustering, particularly the agglomerative
variant~\citep{Rokach2005}, is well-suited for capturing graded relationships
between tasks~\citep{Liu2017}. Unlike flat clustering methods such as
k-means~\citep{Jin2010}, hierarchical clustering builds a tree-like structure
that reflects nested task similarities. This makes it especially useful in
\ac{MTL} settings where task relationships are often complex and nonuniform.

While clustering focuses on grouping tasks to exploit shared structures in
\ac{MTL}, another widely used approach in supervised learning is
\ac{GB}~\citep{Friedman2001}.~\ac{GB} is an ensemble learning approach that has
become one of the most prominent methods in supervised learning, particularly
for tabular datasets~\citep{Bentejac2021,Ravid2022}. Its high performance has
inspired the development of several successful derivatives, including
XGBoost~\citep{Chen2016}, \ac{LGBM}~\citep{Ke2017}, and
CatBoost~\citep{Prokhorenkova2018}. Only a limited number of studies have
investigated extending \ac{GB} to \ac{MTL}, enabling the model to capture both
shared and task-specific patterns simultaneously~\citep{Olivier2011,Emami2023}.
A more robust variant addresses task diversity by clustering tasks, optimizing
model parameters within each cluster, and softly weighting task-cluster
assignments using a sigmoid function~\citep{Emami2025}.

A central challenge in ensemble-based \ac{MTL} is task heterogeneity, where
tasks differ substantially in input distributions, noise levels, or output
structures. When heterogeneous or outlier tasks are forced into joint training,
they can distort shared representations and degrade overall
performance~\citep{Yu2007,Gong2012}. As a result, existing ensemble-based
extensions to \ac{MTL} often lack robustness to such heterogeneity. For
example, \ac{MTGB} extends \ac{GB} to the \ac{MTL} setting by decomposing each
task into shared and task-specific components~\citep{Emami2023}. However,
because all tasks contribute equally to learning the shared structure,
\ac{MTGB} remains sensitive to outlier tasks that deviate strongly from the
majority, which can negatively influence the ensemble~\citep{Emami2025}. To
address this limitation, a robust extension, \ac{R-MTGB}, was proposed
in~\citep{Emami2025}. It incorporates a learnable inlier-outlier mechanism that
assigns extreme weights to anomalous tasks. Although effective at mitigating
the impact of severe outliers, \ac{R-MTGB} relies on a binary assumption in
which tasks are categorized strictly as either inliers or outliers, with
robustness achieved through this twofold separation. This assumption is
restrictive in settings where tasks naturally form multiple heterogeneous
groups. More broadly, although \ac{GB} has been extended to \ac{MTL}, existing
approaches do not perform task clustering in the sense of \ac{CMTL}.
Consequently, the integration of clustering-based \ac{MTL} with boosting
(capable of modeling multiple task groups) remains largely unexplored.

To address these challenges, we introduce \ac{RMB-CLE}, a robust \ac{MTL}
framework grounded in error-based task relatedness. Rather than relying on
in-domain performance or heuristic similarities, \ac{RMB-CLE} estimates
functional task similarity through cross-task generalization errors, which
provide a principled basis for transferability and cross-task risk. These
error-driven representations define a similarity geometry over tasks, from
which latent task groups are discovered via hierarchical agglomerative
clustering with automatic model selection. Learning then proceeds through
cluster-wise local ensembling, enabling selective information sharing within
functionally coherent task groups. By jointly integrating error-driven
similarity estimation, adaptive task clustering, and localized ensemble
learning, \ac{RMB-CLE} structurally mitigates negative transfer under task
heterogeneity, going beyond ad-hoc clustering-boosting combinations.

The main contributions of this work are:
\begin{itemize}
      \item \textbf{A principled framework for robust multi-task learning:}
            We introduce \ac{RMB-CLE}, which infers task relatedness
            from cross-task generalization errors, providing a
            theoretically grounded measure of functional
            similarity that reflects transferability rather
            than task difficulty.
      \item \textbf{Error-driven discovery of latent task structure:}
            By embedding tasks in an error-triggered similarity
            space and applying adaptive hierarchical clustering,
            \ac{RMB-CLE} automatically uncovers heterogeneous
            task groups without metadata, predefined
            partitions, or low-rank assumptions.
      \item \textbf{Mitigation of negative transfer via local ensembling:}
            Learning is performed through cluster-wise ensemble
            models that enable selective information sharing
            within coherent task groups while isolating
            incompatible tasks, resulting in robustness
            under distributional shift and task heterogeneity.
      \item \textbf{Extensive empirical validation:}
            Experiments on synthetic and real-world benchmarks
            demonstrate that \ac{RMB-CLE} accurately recovers latent
            task clusters and consistently outperforms
            single-task, pooling-based, and existing multi-task
            boosting methods.
\end{itemize}

The remainder of this paper is organized as follows.
Section~\ref{sec:related_work} reviews related work.
Section~\ref{sec:methodology} presents the proposed methodology, including the
mathematical framework and theoretical motivation.
Section~\ref{sec:experiments} describes the experimental setup and reports the
results. Section~\ref{sec:conclusions} concludes with the main findings.
Additional ablation studies and supplementary experimental results are provided
in~\ref{appendix:ablation_study} and~\ref{appendix:additional_experiments}.

\section{Related work}~\label{sec:related_work}
One of the earliest studies on clustering in \ac{MTL}
proposed grouping tasks based on similarity to
limit the influence of outliers~\citep{Thrun1996}.
The approach relies on a weighted nearest-neighbor
framework with Euclidean distances to compute a
task transfer matrix that quantifies pairwise relatedness.
Another similar study
proposed an \ac{MTL} approach that incorporates task clustering to enhance the
Bayesian learning process~\citep{bakker2003}. In this framework, neural networks serve as the base
estimators, where the input-to-hidden weights are shared across all tasks,
while the hidden-to-output weights are \ac{TS}. The \ac{TS} parameters are
assumed to be drawn from a Gaussian prior, and the shared parameters are
learned empirically using maximum likelihood estimation (empirical Bayes) over
all tasks. A Bayesian nonparametric framework for \ac{MTL} was introduced later
by~\citep{Xue2007}, employing Dirichlet Process priors to automatically cluster
related tasks. The framework supports both Symmetric MTL (SMTL), which performs
joint learning, and Asymmetric MTL (AMTL), which enables knowledge transfer
without access to the original data. Logistic regression parameters are drawn
from a Dirichlet Process-based prior. Variational Bayesian inference is applied
in SMTL, and the resulting SMTL posterior is used as a prior in AMTL.

Subsequent research introduced convex and subspace-based formulations for
clustering tasks, beginning with the cluster norm of a regularization that
balances within-cluster compactness and between-cluster separation, encouraging
\ac{TS} weight vectors to form a small number of clusters~\cite{Jacob2008}.
Learning the shared subspace for multi-task clustering is proposed
in~\cite{Gu2009}, assuming that despite distributional differences, tasks align
in a latent space. The model jointly clusters within tasks and across tasks,
balanced by a regularization parameter, and optimizes assignments and the
projection matrix via alternating minimization with convergence guarantees. A
similar study addresses task heterogeneity by aligning distributions across
tasks while preserving within-task structure~\cite{Gu2011}. Using Kernel
Hilbert Space with Kernel Mean Matching for alignment and Graph Regularization
with Laplacian constraints for structure, kernel k-means in the learned space
integrates both sources of information for effective transfer. Later, a unified
theoretical framework was proposed in~\cite{Zhou2011}, showing the equivalence
of Alternating Structure Optimization, which assumes a common low-dimensional
feature space, and \ac{CMTL}, which assumes clustered tasks with shared
features. Both are reformulated as equivalent optimization problems through
spectral relaxation. In addition, a convex relaxation of the non-convex
\ac{CMTL} problem is introduced to improve computational efficiency in
high-dimensional settings.

Later methods leveraged kernel learning and spectral techniques for cross-task
clustering. A learning-based formulation was introduced, combining multi-task
feature learning with multiclass maximum-margin clustering~\citep{Zhang2015}.
Building on this, multi-task spectral clustering was developed to exploit
inter-task correlations through a shared low-dimensional representation with
$\ell_{2,p}$-norm regularization, while also preserving intra-task structure,
enabling both cluster label learning and mapping for out-of-sample
data~\citep{Yang2015}. In a related line of work, the Self-Adapted Multi-Task
Clustering model enhances cross-task clustering while reducing negative
transfer by selectively reusing instances, using divergence- and kernel-based
similarity within a spectral clustering framework to adapt dynamically to task
relatedness~\citep{Zhang2016adapted}. Flexible-\ac{CMTL} incorporates
instance-based learning by introducing representative tasks as reference points
for organizing task relationships, where each task connects to multiple
representatives with different weights, allowing flexible soft clustering and
adaptive information sharing; the number of clusters is inferred automatically
from the data~\citep{Zhou2016}. In addition, a model-based clustering approach
performs within-task clustering via symmetric non-negative matrix factorization
with linear regression and learns cross-task relatedness by estimating
similarities of regression weights to enable knowledge transfer, with an
alternating optimization algorithm handling non-convexity~\citep{Zhang2018}.

A closely related line of work in deep \ac{MTL} addresses negative transfer by
manipulating task gradients during optimization rather than relying on static
or heuristic loss weighting. Early work in this direction formulates \ac{MTL}
as a multi-objective optimization problem, where the goal is to identify
Pareto-stationary solutions that balance competing task objectives instead of
minimizing a weighted sum of losses, leading to gradient-based procedures that
compute a common descent direction for all tasks and scalable approximations
suitable for \ac{DNN}~\cite{Sener2018}. Building on this perspective,
subsequent methods focus on resolving gradient conflicts directly at each
optimization step: one approach modifies task gradients through projection
whenever their directions conflict, effectively removing components that would
harm progress on other tasks while remaining model-agnostic and compatible with
standard optimizers~\cite{Yu2020}. More recent work further refines
gradient-based aggregation by introducing constrained optimization formulations
that control the trade-off between improving average task performance and
protecting the worst-performing tasks, resulting more balanced solutions along
the Pareto frontier in deep \ac{MTL}~\cite{Liu2021}.

A complementary line of work addresses negative transfer in \ac{MTL} by
reasoning about relationships between tasks, rather than modifying optimization
dynamics. In the context of transfer learning, Taskonomy studies task
relatedness by measuring how representations learned for one task transfer to
another, yielding a directed task graph that guides supervision and transfer
decisions~\cite{Zamir2018}. However, subsequent work has shown that transfer
affinity does not necessarily predict whether tasks should be learned jointly
in a multi-task setting, as multi-task compatibility depends strongly on
factors such as model capacity, dataset size, and optimization interactions
~\cite{Standley2020}.

A recent line of research has extended boosting methods to \ac{MTL} setting. In
particular, a two-phase framework known as \ac{MTGB} has been proposed to
combine shared and task-specific learning within boosting~\cite{Emami2023}. In
the shared learning phase, \ac{MTGB} fits common base learners across all tasks
to extract representations that capture global structure. In the second phase
(task-specific learning), it augments these shared representations with \ac{TS}
learners trained on the corresponding pseudo-residuals, thereby developing the
model to individual tasks. This design allows \ac{MTGB} to balance knowledge
transfer with specialization. Despite these advantages, \ac{MTGB} assumes that
all tasks should contribute uniformly to the global representation, making it
vulnerable to negative transfer in the presence of heterogeneous or noisy
tasks~\citep{Emami2025}. Its extension, \ac{R-MTGB}, addresses this issue by
adding an intermediate block that introduces a learnable sigmoid-based
weighting scheme~\citep{Emami2025}. This mechanism softly partitions tasks into
outliers and non-outliers, down-weighting adversarial ones before proceeding to
task-specific fine-tuning. While \ac{R-MTGB} improves robustness, its binary
partitioning restricts flexibility: it can only distinguish between two groups
of tasks, ignoring more complex cluster structures.

\subsection{Comparative analysis}~\label{subsec:comparison}
Our proposed \ac{RMB-CLE} framework advances \ac{CMTL} by utilizing a
principled, unsupervised agglomerative clustering method to discover latent
task relationships through the alignment of cross-task errors. Unlike
conventional \ac{CMTL} methods that rely on predefined clusters or low-rank
projections, Bayesian nonparametric models~\citep{Xue2007} requiring complex
inference, convex regularization approaches such as cluster
norm~\citep{Jacob2008}, or subspace learning methods~\citep{Gu2009,Gu2011}
assuming global alignment, \ac{RMB-CLE} derives similarity directly from
error-based inter-task behavior, leveraging task-specific local ensembles to
remain robust to distributional shifts and heterogeneity. In contrast to
spectral clustering~\citep{Yang2015}, or
early transfer approaches~\citep{Thrun1996}, \ac{RMB-CLE} leverages cross-task
error signals to construct adaptive cluster-level ensembles. It remains
model-agnostic and avoids reliance on metadata, handcrafted similarity
measures, or nearest-neighbor assumptions.

In contrast to gradient-based deep \ac{MTL} methods, which mitigate negative
transfer by adjusting gradients within a shared
parameterization~\cite{Sener2018, Yu2020, Liu2021}, \ac{RMB-CLE} operates at
the level of task relationships. Gradient-based approaches assume that all
tasks remain coupled throughout training and address incompatibilities only
through local, synchronous optimization corrections. As a result, they neither
uncover latent task clusters nor enforce structural separation between
incompatible tasks.~\ac{RMB-CLE} instead infers functional similarity from
cross-task generalization errors and partitions tasks into clusters, preventing
negative transfer by isolating unrelated tasks rather than attenuating gradient
interference within a shared model.

Complementary to representation-based transfer analyses~\cite{Zamir2018}, and
model-dependent task grouping strategies~\cite{Standley2020}, our approach
defines task similarity directly through cross-task predictive performance and
leverages this structure for robust clustered learning. This formulation is
model-agnostic, applies beyond deep neural networks, and directly targets the
reduction of negative transfer in heterogeneous multi-task settings.

Compared to recent boosting-based approaches such as
\ac{MTGB}~\citep{Emami2023} and its robust extension
\ac{R-MTGB}~\citep{Emami2025}, our proposed approach removes the assumption of
a fixed or binary partition of tasks. These limitations highlight the need for
a more general approach that can adaptively discover multiple task clusters,
safeguard against negative transfer without relying on hard or binary
partitions, and still preserve task-specific specialization.

Table~\ref{tab:sota_mtl} summarizes the key methodological distinctions among
representative \ac{MTL} approaches. In particular, it highlights that
gradient-based deep \ac{MTL} methods mitigate negative transfer through
optimization-level interventions within a fully shared model, without modeling
task structure. Boosting-based multi-task methods therefore form the most
relevant comparison class for \ac{RMB-CLE}. Beyond their empirical
effectiveness on tabular data~\cite{Ravid2022}, they share the ensemble-based
learning paradigm underlying \ac{RMB-CLE} and enable controlled information
sharing without relying on shared deep representations. Within this class,
existing approaches such as \ac{MTGB} and \ac{R-MTGB} enhance robustness via
staged boosting or outlier-aware weighting, but remain restricted to global or
binary cluster structures and do not support adaptive multi-cluster discovery
or cluster-level ensembling.

\begin{table}[t]
      \centering
      \caption{Comparison of representative multi-task learning methods.
            \textbf{Grad}: gradient manipulation;
            \textbf{Clust}: discovers task clusters; \textbf{Ens}: uses cluster-level ensembles;
            \textbf{Meta}: requires metadata; \textbf{Agn}: model-agnostic.
            (Symbols: \cmark=present; \xmark=absent).}
      \small
      \setlength{\tabcolsep}{4pt}
      \renewcommand{\arraystretch}{1.15}
      \begin{tabularx}{\linewidth}{@{}l *{5}{>{\centering\arraybackslash}X}@{}}
            \toprule
            \textbf{Method}                           & \textbf{Grad} & \textbf{Clust} & \textbf{Ens} & \textbf{Meta} & \textbf{Agn} \\
            \midrule
            Bayesian-MTL~\citep{Xue2007}              & \xmark        & \cmark         & \xmark       & \xmark        & \xmark       \\
            Cluster norm~\citep{Jacob2008}            & \xmark        & \cmark         & \xmark       & \xmark        & \xmark       \\
            Shared subspace MTL~\citep{Gu2009,Gu2011} & \xmark        & \xmark         & \xmark       & \xmark        & \xmark       \\
            Spectral task clustering~\citep{Yang2015} & \xmark        & \cmark         & \xmark       & \xmark        & \xmark       \\
            Gradient-based deep MTL~\citep{Sener2018} & \cmark        & \xmark         & \xmark       & \xmark        & \cmark       \\
            Gradient surgery~\citep{Yu2020}           & \cmark        & \xmark         & \xmark       & \xmark        & \cmark       \\
            Conflict-averse gradients~\citep{Liu2021} & \cmark        & \xmark         & \xmark       & \xmark        & \cmark       \\
            MTGB~\citep{Emami2023}                    & \xmark        & \xmark         & \xmark       & \xmark        & \cmark       \\
            R-MTGB~\citep{Emami2025}                  & \xmark        & \cmark         & \xmark       & \xmark        & \cmark       \\
            \midrule
            RMB-CLE (ours)                            & \xmark        & \cmark         & \cmark       & \xmark        & \cmark       \\
            \bottomrule
      \end{tabularx}

      \vspace{2pt}
      \label{tab:sota_mtl}
\end{table}

\section{Methodology}~\label{sec:methodology}

In this work, \ac{RMB-CLE} is proposed as a framework for identifying and
clustering related tasks and training local ensembles within each cluster,
enabling knowledge sharing while isolating unrelated tasks to mitigate negative
transfer. This section first introduces the necessary preliminaries
(Subsection~\ref{subsec:preliminaries}), then presents the mathematical
formulation and algorithmic details of the framework
(Subsection~\ref{subsec:framework}), and finally provides its theoretical
justification (Subsection~\ref{subsec:theoretical_motivation}).

\subsection{Preliminaries}~\label{subsec:preliminaries}
Let $\mathcal{T}=\{1,\dots,m\}$ denote the set of supervised tasks. Each task
$i\in\mathcal{T}$ is associated with a labeled dataset
$\mathcal{D}_i=\{(\mathbf{x}_{i,v},y_{i,v})\}_{v=1}^{n_i}$, where
$\mathbf{x}_{i,v}\in\mathbb{R}^d$ are input features and $y_{i,v}\in\mathcal{Y}$
denotes the corresponding supervised target.

For each task $i$, we first fit a predictor $F_i$ via empirical risk
minimization,
\begin{equation}
      F_i = \argmin_{F\in\mathcal{H}}
      \frac{1}{n_i}\sum_{v=1}^{n_i}\ell\big(y_{i,v},F(\mathbf{x}_{i,v})\big),
      \label{eq:task-fit}
\end{equation}
where $\ell$ is the task-appropriate loss and $\mathcal{H}$ denotes a
hypothesis space of predictive functions.
For classification tasks, $\ell$ is given by the cross-entropy loss,
\begin{equation}~\label{eq:CE}
      \ell(y_{i,v}, F_i) =
      -\sum_{q=1}^{Q} y_{i,q} \ln(P_q),
\end{equation}
where $Q$ denotes the number of class labels, $y_{i,q} \in \{0,1\}$ is the
$q$-th component of the one-hot encoded target vector, and $P_{i,q}$ is the
predicted probability associated with class $q$, given by
$P_{i,q} = \exp(F_{i,q})\big/\sum_{q=1}^{Q}\exp(F_{i,q})$.
For regression tasks, we instead use the squared error loss,
\begin{equation}~\label{eq:SE}
      \ell(y_{i,v},F_i) = \frac{1}{2}{(y_{i,v} - F_i)}^2.
\end{equation}
In practice, we instantiate $\mathcal{H}$ as the class of
\ac{LGBM}~\citep{Ke2017} ensembles ($\mathcal{H}_{\text{LGBM}}$), using
decision tree regressor with maximum depth of one (decision stumps)
($h_{i,t}$), trained iteratively with learning rate $\nu \in (0, 1]$,
\begin{equation}
      F_{i, t}(\mathbf{x})=F_{i, (t-1)}(\mathbf{x})
      +\nu\,h_{i,t}(\mathbf{x}),
      \label{eq:lgbm-recursion}
\end{equation}
where $t = 1,\ldots,T$ indexes the boosting iterations
and $T$ denotes the total number of boosting rounds.
In the remainder of this section, we denote by $F_i$ the final
boosted predictor obtained after $T$ boosting iterations,
i.e., $F_i \equiv F_{i,T}$.

We also introduce the notion of a boosting block. A boosting block
$\mathcal{S}_{(.)}$ denotes a structural unit composed of a fixed number of
sequential boosting iterations (i.e., base learners) trained under a specific
functional role. Different stages may serve different purposes within a model,
such as learning shared representations ($\mathcal{S}_{(1)}$), task-specific
components ($\mathcal{S}_{(2)}$), outlier-aware weighting
($\mathcal{S}_{(3)}$), or cluster-level ensembles ($\mathcal{S}_{(4)}$).
Accordingly, the size of each block, denoted by
$\mathcal{K}_{\mathcal{S}_{(.)}}$, may vary across blocks depending on their
role in the model.

\subsection{Framework formulation}~\label{subsec:framework}
\paragraph{Step 1: Cross-Task Similarity Estimation}
To calculate task similarity, we first quantify how well model $F_j$ (trained
on task $j$) transfers to task $i$ by evaluating it on $\mathcal{D}_i$,
\begin{equation}
      E_{i,j} =
      \begin{cases}
            \tfrac{1}{n_i}\displaystyle\sum_{v=1}^{n_i}\big(y_{i,v}-F_j(\mathbf{x}_{i,v})\big)^2,
             & \text{Regression},     \\[1ex]
            1-\tfrac{1}{n_i}\displaystyle\sum_{v=1}^{n_i}
            \mathbf{1}\!\left\{F_j(\mathbf{x}_{i,v})=y_{i,v}\right\},
             & \text{Classification},
      \end{cases}
      \label{eq:cross-error}
\end{equation}
where, $\mathbf{1}\{\cdot\}$
is the indicator function.

\medskip
\noindent
\textbf{Remark.}
Note that cross-task evaluation is essential,
since in-domain accuracy only reflects task difficulty
(e.g., noise level or sample size) and not inter-task
compatibility. Two unrelated tasks may both be \emph{easy}
(low in-domain error) but transfer poorly, while two
\emph{hard} tasks may transfer well if their conditional
mechanisms are similar. By measuring
transferability between tasks, we avoid negative
knowledge transfer (see Subsection~\ref{subsec:theoretical_motivation}) and obtain a principled similarity
structure. As a potential alternative, one could consider
an out-of-sample cross-task error approach, which we
discuss and contrast in~\ref{appendix:alternative_approach}.
\medskip

We then convert the errors to similarities, where smaller cross-task error
indicates greater closeness,
\begin{equation}
      \mathbf{s}_{i,j}=\frac{1}{E_{i,j}+\varepsilon},\qquad \varepsilon>0.
      \label{eq:sim}
\end{equation}
Collecting these entries across all task pairs returns the
similarity matrix,
\begin{equation}
      \mathbf{S} = (\mathbf{s}_{i,j})_{i,j=1}^m \in\mathbb{R}^{m\times m}.
      \label{eq:similarity_matrix}
\end{equation}
Thus, each vector $\mathbf{s}_i = (s_{i,1},\dots,s_{i,m})$ array can be viewed
as a \emph{similarity profile} of task $i$ relative to all other tasks.
In this sense, every task is embedded as a vector in
$\mathbb{R}^m$, and tasks with similar predictive behavior
correspond to similarity vectors that are close in angle.
This motivates the use of cosine distance,
\begin{equation}
      \Delta_{i,j}=1-\frac{\langle \mathbf{s}_i,\mathbf{s}_j
            \rangle}{\|\mathbf{s}_i\|_2\,\|\mathbf{s}_j\|_2},\quad
      \Delta_{i,j}\in[0,1].
      \label{eq:cosdist}
\end{equation}

\paragraph{Step 2: Task clustering}
For clustering, we define the average linkage distance between two multi-task
clusters $A$ and $B$,
\begin{equation}
      \delta(A,B)=\frac{1}{|A|\,|B|}\sum_{i\in A}\sum_{j\in B} \Delta_{i,j}.
      \label{eq:avg-link}
\end{equation}
Applying agglomerative clustering with average linkage (also known as
Unweighted Pair Group Method with Arithmetic mean (UPGMA))~\cite{Sokal1985}
on $\Delta$
produces a dendrogram $\mathcal{L}$ with $m$ leaves.
For each $k \in \{2,\dots,\min(m,K_{\max})\}$,
we obtain a partition (assignment function for a given $k$)
$g_k:\{1,\dots,m\}\to\{1,\dots,k\}$ with clusters
$\mathcal{C}_c = \{i : g_k(i)=c\}$.

\paragraph{Step 3: Optimal Cluster Search}
To choose the best number of clusters, we evaluate the silhouette
score~\citep{Peter1987},
\begin{equation}
      \varphi(i) = \frac{b(i)-a(i)}{\max\{a(i),b(i)\}},                         \\[0.5em]
      \label{eq:silhouette}
\end{equation}
\noindent where $a(i)=\frac{1}{|C_{g_k(i)}|-1}
      \sum_{j\in \mathcal{C}_{g_k(i)}\setminus\{i\}} \Delta_{i,j}$
is the inner-cluster distance,
and $b(i)=\min_{c\neq g_k(i)}
      \frac{1}{|\mathcal{C}_c|}\sum_{j\in \mathcal{C}_c} \Delta_{i,j}$
is the nearest other-cluster distance.
Hence the optimal cluster count is,
\begin{equation}
      k^\star = \argmax_{k}\,\frac{1}{m}\sum_{i=1}^m \varphi(i),
      \label{eq:bestk}
\end{equation}
and the final assignment will be computed as $g(i)=g_{k^\star}(i)$.

\paragraph{Step 4: Agglomerative update}
Using the Lance and Williams recurrence~\cite{Murtagh2017}, when clusters $A$
and $B$ merge, their distance ($\delta$) to any other cluster $C$ is updated
as,
\[
      \begin{aligned}
            \delta(A\cup B, C) & = \alpha_A \delta(A,C)+\alpha_B \delta(B,C) \\
                               & \quad + \beta
            \delta(A,B)+\gamma\,|\delta(A,C)-\delta(B,C)|.
      \end{aligned}
\]
For average linkage (UPGMA), the parameters are given by $\alpha_A =
      \tfrac{|A|}{|A|+|B|}$, $\alpha_B = \tfrac{|B|}{|A|+|B|}$, and $\beta = \gamma =
      0$. The distance update rule is,
\begin{equation}
      \delta(A \cup B, C) \;=\; \frac{|A|\,\delta(A,C) +
            |B|\,\delta(B,C)}{|A|+|B|},
\end{equation}
which ensures that after each merge,
cluster-to-cluster distances are updated consistently,
enabling the construction of the full dendrogram $\mathcal{L}$.

\paragraph{Step 5: Cluster-specific models}
Finally, for each cluster $c$, we pool data $\mathcal{U}_c=\bigcup_{i\in
            C_c}\mathcal{D}_i$ and train a specialized local ensemble $f$,
\begin{equation}
      f_c=\argmin_{f\in\mathcal{F}}\
      \tfrac{1}{|\mathcal{U}_c|}\sum_{(\mathbf{x},y,i)\in\mathcal{U}_c}
      \ell\big(y, f([\mathbf{x};i])\big),
      \label{eq:local-ensemble}
\end{equation}
where $\ell$ is the loss function, defined as the
cross-entropy loss (Eq.~\eqref{eq:CE}) for classification tasks and the squared
error loss (Eq.~\eqref{eq:SE}) for regression tasks.
In this study we instantiate
$\mathcal{F}$ with \ac{LGBM}~\citep{Ke2017},
which introduces histogram-based feature binning
and leaf-wise (Best-first) tree growth for fast,
memory-efficient boosting with depth equal to one
(decision stumps)
and \ac{MTGB}~\citep{Emami2023}, which first
learns a shared latent function across tasks
to capture common structure and then task-specific
functions to model individual particularities,
both using decision stumps as base learners.
Other local \ac{ML} models are also compatible.

\paragraph{Prediction}
At prediction time, each input instance comes with a task identifier $i \in
      \mathcal{T}$. Since cluster assignments were already determined during
training, prediction reduces to a cluster lookup followed by cluster-specific
inference,
\begin{equation}
      c = g(i),
      \qquad
      \hat{y} = f_{c}\big([\mathbf{x};i]\big),
\end{equation}
where \(\hat{y}\) is the predicted output for input \(\mathbf{x}\) under task
$i$, $g:\mathcal{T}\to\{1,\dots,k^\star\}$ is the final task-to-cluster
assignment and $f_c$ is the local ensemble trained on $c$.

For clarity, we detail the algorithmic structure of \ac{RMB-CLE} and provide
its pseudocode in Algorithm~\ref{alg:rmb_cle}.

\begin{algorithm}[tbp]
      \footnotesize
      \caption{RMB-CLE Training Algorithm}
      \label{alg:rmb_cle}
      \begin{algorithmic}[1]
            \Require $\{\mathcal{D}_i\}_{i=1}^m$, $\varepsilon>0, \text{local ensemble} f \in \mathcal{F}$
            \State \textbf{Apply LGBM:} \For{$i=1..m$} $F_i \gets \text{TrainLGBM}(\mathcal{D}_i)$ \EndFor
            \State \textbf{Cross-task errors \& similarities:}
            \For{$i=1..m$}\For{$j=1..m$}
            \State $E_{i,j} \gets$ Eq.~\eqref{eq:cross-error}; \quad $\mathbf{s}_{i,j}\gets 1/(E_{i,j}+\varepsilon)$
            \EndFor\EndFor
            \State \textbf{Cosine distances on similarity arrays:}
            \For{$i,j=1..m$}
            \State $\Delta_{i,j}\gets \begin{cases}
                        0,                                                                                       & i=j     \\
                        1-\frac{\langle \mathbf{s}_i,\mathbf{s}_j\rangle}{\|\mathbf{s}_i\|_2\|\mathbf{s}_j\|_2}, & i\neq j
                  \end{cases}$
            \EndFor \State $\Delta \gets \tfrac12(\Delta+\Delta^\top)$
            \State \textbf{Agglomerative clustering (UPGMA):}
            \State $Z \gets \texttt{linkage}(\texttt{squareform}(\Delta))$
            \For{$k=2..\min(m,K_{\max})$}
            \State $g_k \gets \texttt{fcluster}(Z)$
            \State \textbf{Silhouette at $k$:}
            \For{$i=1..m$}
            \State $A \gets \{j: g_k(j)=g_k(i)\}$
            \State $a(i)\gets \begin{cases}0,& |A|=1\\ \frac{1}{|A|-1}\sum_{j\in A\setminus\{i\}}\Delta_{i,j},&\text{else}\end{cases}$
            \State $b(i)\gets \min\limits_{c\neq g_k(i)} \frac{1}{|C_c|}\sum_{j\in \mathcal{C}_c} \Delta_{i,j}$ where $\mathcal{C}_c{=}\{j:g_k(j){=}c\}$
            \State $k^\star \gets \argmax_{k} \varphi_k(i)$ {Eq.~\eqref{eq:silhouette}}
            \EndFor
            \State $\bar\varphi(k)\gets \frac{1}{m}\sum_{i=1}^m \varphi_k(i)$
            \EndFor
            \State $k^\star \gets \arg\max_k \bar\varphi(k)$; \ $g(i)\gets g_{k^\star}(i)$; \ $\mathcal{C}\gets\{i:g(i)=c\}$
            \State \textbf{local ensembles:} \For{each $c$}
            \State $\mathcal{U}_c\gets \bigcup_{i\in C_c}\mathcal{D}_i$; \
            $f_c \gets f_c\{([\mathbf{x};i],y):(\mathbf{x},y,i)\in\mathcal{U}_c\}$
            \EndFor
            \State \textbf{return } \textsc{RMB-CLE-via-$\mathcal{F}$}
      \end{algorithmic}
\end{algorithm}

\subsection{Theoretical motivation}~\label{subsec:theoretical_motivation}
This section justifies why cross-task evaluation (Eq.~\eqref{eq:cross-error})
provides a principled measure of task similarity, whereas evaluating each model
only on its own dataset does not.

The key theoretical insight underlying \ac{RMB-CLE} is that cross-task
generalization error provides a principled basis for task compatibility. In
both regression and classification settings, the risk arising when transferring
a model across tasks decomposes into a task-dependent noise term and a
functional mismatch term that reflects differences between the underlying
conditional mechanisms. Consequently, cross-task errors capture functional
relatedness rather than task difficulty or noise alone.

By clustering tasks based on this decomposition, \ac{RMB-CLE} avoids forcing
information sharing between incompatible tasks. Tasks that exhibit large
functional mismatch are separated into different clusters, while tasks with
compatible conditional mechanisms are grouped together, enabling selective
knowledge sharing. This provides a safeguard against negative multi-task
transfer.

We formalize this argument in two settings. The regression analysis in
Subsection~\ref{subsubsec:theory_reg} shows that cross-task risk admits an
exact decomposition into a functional distance term and an irreducible noise
term. The classification analysis in Subsection~\ref{subsubsec:theory_clf}
establishes that cross-task error upper-bounds the excess risk in terms of
disagreement with the Bayes classifier.

\subsubsection{Regression setting}~\label{subsubsec:theory_reg}

Let task $i$ be associated with joint distribution $P_i(X,Y)$, where $X \in
      \mathcal{X}$ denotes inputs and $Y \in \mathbb{R}$ the output. The regression
function is defined as,
\begin{equation}
      \eta_i(x) = \mathbb{E}[Y \mid X=\mathbf{x}, i],
\end{equation}
that is, the conditional expectation of $Y$ given $X=\mathbf{x}$ for task $i$.
We also define the conditional noise variance as,
\begin{equation}
      \sigma_i^2(\mathbf{x}) = \mathbb{E}\!\left[(Y-\eta_i(\mathbf{x}))^2 \mid X=\mathbf{x}, i\right].
\end{equation}

Considering $F_j$ as the predictor trained on task $j$. The squared-loss risk
of $F_j$ on task $i$ is
\begin{equation}
      \mathcal{R}_{i}(F_j) = \mathbb{E}_{P_i}\!\big[(Y-F_j(X))^2\big],
\end{equation}
where ${P_i}$ denotes the joint data-generating distribution of $(X,Y)$ for
task $i$.

\paragraph{Step 1: Conditioning on $X$}
Using the law of total expectation,
\begin{equation}
      \mathcal{R}_{i}(F_j)
      = \mathbb{E}_{P_i^X}\!
      \Big[ \mathbb{E}\!\big[(Y-F_j(X))^2 \mid X\big] \Big],
\end{equation}
where $P_i^X$ denotes the marginal distribution of the inputs $X$ under task
$i$, obtained from the joint distribution $P_i(X,Y)$.
The inner expectation is taken with respect to the conditional distribution
$P_i^{Y\mid X}$, i.e., $P_i(Y\mid X)$.

\paragraph{Step 2: Expanding inside the conditional expectation}
For a fixed $X=\mathbf{x}$ we write $Y = \eta_i(\mathbf{x}) + \varepsilon_i$,
where $\mathbb{E}[\varepsilon_i \mid X=\mathbf{x}]=0$ and
$\mathbb{E}[\varepsilon_i^2 \mid X=\mathbf{x}] = \sigma_i^2(\mathbf{x})$. Then,
\begin{equation}
      \mathbb{E}[(Y-F_j(x))^2 \mid X=x]
      = \mathbb{E}\!\left[(\eta_i(\mathbf{x})
            + \varepsilon_i - F_j(\mathbf{x}))^2 \mid X=\mathbf{x}\right].
\end{equation}
Expanding the square gives,
\begin{equation}
      \left(\eta_i(\mathbf{x})-F_j(\mathbf{x})\right)^2
      + 2\left(\eta_i(\mathbf{x})-F_j(\mathbf{x})\right)\mathbb{E}[\varepsilon_i\mid X=x]
      + \mathbb{E}[\varepsilon_i^2 \mid X=\mathbf{x}].
\end{equation}
The middle term vanishes because $\mathbb{E}[\varepsilon_i \mid X=x]=0$, hence,
\begin{equation}
      \mathbb{E}[(Y-F_j(\mathbf{x}))^2 \mid X=\mathbf{x}]
      = (\eta_i(x)-F_j(\mathbf{x}))^2 + \sigma_i^2(\mathbf{x}).
\end{equation}

\paragraph{Step 3: Taking expectation over $X$}
Plugging this back gives,
\begin{equation}
      \mathcal{R}_{i}(F_j)
      = \mathbb{E}_{P_i^X}\!\big[(\eta_i(X)-F_j(X))^2 + \sigma_i^2(X)\big].
\end{equation}
Separating the terms,
\begin{equation}
      \mathcal{R}_{i}(F_j)
      = \underbrace{\mathbb{E}_{P_i}\!\big[(\eta_i(X)-F_j(X))^2\big]}_{\mcirc{1}}
      + \underbrace{\mathbb{E}_{P_i}\!\big[\sigma_i^2(X)\big]}_{\mcirc{2}}.
\end{equation}

Thus, the cross-task error decomposes into $\mcirc{1}$ a task similarity term
measuring the squared $L^2$-distance between the conditional functions $\eta_i$
and $F_j$, and $\mcirc{2}$ a noise term that is independent of $F_j$.

\subsubsection{Classification setting}~\label{subsubsec:theory_clf}

For each task $i \in \{1, \ldots, m\}$, let $(X, Y) \sim P_i$, where $X \in
      \mathcal{X}$ denotes the feature vector and $Y \in \{1,\ldots,K\}$ the class
label, with $K\ge 2$. We write $\Pr(\cdot)$ for probability with respect to the
task-specific distribution $P_i$. The class-posterior function for task $i$ is
defined as
\begin{equation}
      {\eta}_i(x)
      =
      \big(\Pr(Y=1\mid X=\mathbf{x},i),\ldots,\Pr(Y=K\mid X=\mathbf{x},i)\big).
\end{equation}
The corresponding Bayes classifier is
\begin{equation}
      b_i(\mathbf{x})
      =
      \argmax_{k\in\{1,\ldots,K\}} \Pr(Y=k\mid X=\mathbf{x},i).
\end{equation}
For $K=2$, this reduces to the binary Bayes rule
$b_i(\mathbf{x})=\mathbf{1}\{\eta_i(\mathbf{x})\ge \tfrac{1}{2}\}$. For any
classifier $F$, the $(0$--$1)$ risk on task $i$ is
\begin{equation}
      \mathcal{R}_i(F)
      =
      \Pr_{(X,Y)\sim P_i}\big(F(X)\neq Y\big).
\end{equation}

\paragraph{Step 1: Excess risk vs.\ disagreement}
The additional error of using $F_j$ (trained on task $j$) instead of $b_i$
(Bayes on task $i$) satisfies
\begin{equation}
      \mathcal{R}_i(F_j) - \mathcal{R}_i(b_i)
      \le
      \Pr_{P_i^X}\big(F_j(X)\neq b_i(X)\big).
\end{equation}
\emph{Justification.}
Condition on $X=\mathbf{x}$. For a fixed $\mathbf{x}$, the conditional $0$--$1$
risk of a classifier $F$ is $\Pr(Y\neq F(X)\mid X=\mathbf{x})$. The Bayes
classifier $b_i$ minimizes this quantity over all classifiers, hence,
\begin{equation}
      \Pr(Y\neq F_j(X)\mid X=\mathbf{x})
      -
      \Pr(Y\neq b_i(X)\mid X=\mathbf{x})
      \ge 0.
\end{equation}

Moreover, if $F_j(\mathbf{x}) = b_i(\mathbf{x})$, then both classifiers make
the same prediction at $\mathbf{x}$ and therefore have identical conditional
risk, implying that the difference is zero. If $F_j(\mathbf{x}) \neq
      b_i(\mathbf{x})$, then the difference in conditional risks is at most $1$,
since probabilities are bounded by $[0,1]$. Combining both cases returns
\begin{equation}
      \Pr(Y\neq F_j(X)\mid X=\mathbf{x})
      -
      \Pr(Y\neq b_i(X)\mid X=\mathbf{x})
      \le
      \mathbf{1}\{F_j(\mathbf{x})\neq b_i(\mathbf{x})\}.
\end{equation}
Taking expectation with respect to $X \sim P_i^X$ on both sides and using the
law of total expectation,
\begin{equation}\label{eq:risk-bound}
      \begin{aligned}
            \mathcal{R}_i(F_j) - \mathcal{R}_i(b_i)
             & =
            \mathbb{E}_{X\sim P_i^X}\!\left[
                  \Pr(Y\neq F_j(X)\mid X)
                  -
                  \Pr(Y\neq b_i(X)\mid X)
            \right] \\
             & \le
            \mathbb{E}_{X\sim P_i^X}\!\left[
                  \mathbf{1}\{F_j(X)\neq b_i(X)\}
            \right] \\
             & =
            \Pr_{P_i^X}\big(F_j(X)\neq b_i(X)\big).
      \end{aligned}
\end{equation}

\paragraph{Step 2: Cross-task error as a similarity measure}
The bound in Step~1 shows that transferring a model $F_j$ to task $i$ leads to
excess error controlled by the probability of disagreement with the Bayes
classifier of task $i$. When $F_j$ is a statistically consistent approximation
of the latent Bayes rule $b_j$, this disagreement reflects a functional
mismatch between the class-posterior distributions ${\eta}_i$ and ${\eta}_j$.
Therefore, the empirical cross-task error $E_{i,j}$ in
Eq.~\eqref{eq:cross-error} serves as a practical basis for inter-task
compatibility in classification.

To further strengthen the theoretical analysis, we examine the computational
costs of both the proposed and existing models with respect to training and
latency, as detailed in~\ref{apendix:time_cost}. While
the~\ref{apendix:time_cost} shows that the proposed model has higher training
complexity (due to cross-task evaluation and clustering), its prediction is
comparable to or lower than that of the studied models. This efficiency arises
because predictions rely solely on the relevant cluster ensemble, making the
model more time-efficient at test time compared to state-of-the-art multi-task
boosting approaches.

\section{Experiments and results}~\label{sec:experiments}
The proposed approach is evaluated on both synthetic and real-world datasets to
assess performance, robustness, and generalization, and is compared against
state-of-the-art \ac{ST} and \ac{MT} baselines.
Subsection~\ref{subsec:datasets} introduces the datasets used in the
experiments, while Subsection~\ref{subsec:setup} describes the experimental
setup and the considered baseline methods. Results on synthetic and real-world
data are reported in Subsections~\ref{subsec:synthetic}
and~\ref{subsec:realworld}, respectively. Ablation experiments and additional
analyses are presented in~\ref{appendix:ablation_study} and~\ref{appendix:additional_experiments}.

\subsection{Datasets}~\label{subsec:datasets}
To precisely evaluate the performance of the proposed framework, and to test
the robustness of our approach, we employ a diverse set of datasets comprising
both synthetic and real-world \ac{MT} data.

\paragraph{Synthetic datasets}
We define a generic Random Fourier Feature~\citep{Rahimi2007} labeling
function,
\begin{equation}
      \label{eq:rff}
      \Upsilon_{\psi}(\mathbf{x})
      =\sqrt{\frac{2\tau}{\kappa}}
      \sum_{r=1}^{\kappa}\phi_{r}\,
      \cos\!\left(\frac{1}{\lambda d}\,\mathbf{w}_{r}^\top \mathbf{x}+b_{r}\right),
\end{equation}
where the parameter set
\(\psi=(\boldsymbol{\phi},\mathbf{W},\mathbf{b})\) consists of
\(\boldsymbol{\phi}=\{\phi_r\}_{r=1}^{\kappa}\),
\(\mathbf{W}=\{\mathbf{w}_r\}_{r=1}^{\kappa}\) with
\(\mathbf{w}_r \overset{iid}{\sim} \mathcal{N}(\mathbf{0},\mathbf{I}_d)\),
and phase shifts
\(\mathbf{b}=\{b_r\}_{r=1}^{\kappa}\) with
\(b_r \overset{iid}{\sim} \mathcal{U}(0,2\pi)\).
The weights satisfy \(\phi_r \overset{iid}{\sim} \mathcal{N}(0,1)\).
The constants \(\kappa\) and \(d\) denote the number of random features and the
input dimension, respectively, while \(\tau\) is a scaling hyperparameter.
And \(\lambda>0\) denotes the length scale.

This construction corresponds to an approximate draw from a Gaussian
process~\citep{williams2006gaussian} with a shift-invariant squared exponential
kernel.

Building on this foundation, we generate a set of \(\mathcal{C}\) clusters.
Each cluster \(c \in \mathcal{C}\) contains \(m_c\) tasks indexed by \(i \in
\{1,\dots,m_c\}\). For each task \(i\), we define a synthetic labeling function
as the aggregation of a common component (${\text{CC}}$) and a task-specific
(${\text{TS}}$) component. Specifically, we instantiate two independent Random
Fourier Feature functions from~\eqref{eq:rff},
\begin{align}
      \Upsilon_{c}^{\text{CC}}(\mathbf{x}) \equiv \Upsilon_{\psi_{c}^{\text{CC}}}(\mathbf{x}_{i,v}), \\
      \Upsilon_{c}^{\text{TS}}(\mathbf{x}) \equiv \Upsilon_{\psi_{c}^{\text{TS}}}(\mathbf{x}_{i,v}),
\end{align}
where \(\psi_{c}^{\text{CC}}\) and \(\psi_{c}^{\text{TS}}\) are independently
sampled parameter sets. The resulting function values for input
\(\mathbf{x}_i\) are then given by
\begin{align}
      \upsilon^{\text{CC}} & = \Upsilon_{c}^{\text{CC}}(\mathbf{x}_{i,v}), \\
      \upsilon^{\text{TS}} & = \Upsilon_{c}^{\text{TS}}(\mathbf{x}_{i,v}).
\end{align}

Finally, the output for task $i$ in cluster $c$ is generated as a convex
combination of its common and task-specific components,
\begin{equation}
      \label{eq:output_generation}
      y_{i,v}(\mathbf{x}_{i,v}) =
      \omega \,\upsilon^{\text{CC}}(\mathbf{x}_{i,v})
      + (1-\omega)\,\upsilon^{\text{TS}}(\mathbf{x}_{i,v}),
\end{equation}
where $\omega \in (0,1]$ controls the balance between
common and task-specific contributions.
For regression tasks, Eq.~\eqref{eq:output_generation} is used directly,
while for classification tasks, binary labels are obtained by
thresholding Eq.~\eqref{eq:output_generation} at its task-specific median value,
assigning label $1$ to samples with $y_{i,v}$
above the median and $0$ otherwise.

\begin{figure}[ht]
      \centering
      \includegraphics[width=0.7\linewidth]{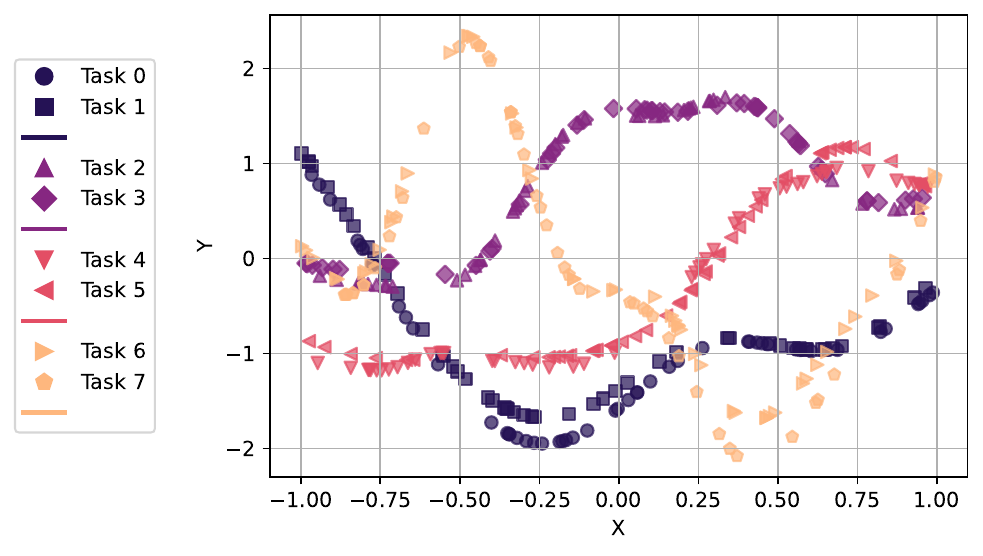}
      \caption{One-dimensional synthetic multi-task dataset with $\mathcal{T}=8$,
            $C=4$, and $\omega = 0.9$. Colors indicate clusters and markers denote
            tasks.}
      \label{fig:Figure_1}
\end{figure}

Figure~\ref{fig:Figure_1} provides a visual example of a generated synthetic
dataset with $\mathcal{T}=8$ tasks in one dimension ($d=1$), grouped into
$\mathcal{C}=4$ clusters, where $\omega = 0.9$ is used. Colors denote clusters,
while symbols distinguish tasks. This illustration is provided for
visualization purposes only.

For the experiments, we generate synthetic data using 100 repetitions with
different random seeds. Each repetition contains 25 tasks evenly distributed
across five clusters, constructed with $\omega=0.9$. Inputs $\mathbf{x}$ are
sampled uniformly from $[-1,1]^{d_x}$, where $d_x$ is the input dimension. Each
task has 300 training samples and 1{,}000 test samples, with five input
features. In regression setting, we use the real-valued outputs from the
data-generating process. In classification setting, continuous outputs are
first generated and converted into binary labels by thresholding at the
task-specific median, computed prior to the train-test split. Samples above the
median are labeled 1 and the rest 0. All generated synthetic datasets are
publicly available via Mendeley
Data\footnote{\href{https://data.mendeley.com/datasets/3gms2fvs93} {
            data.mendeley.com/datasets/3gms2fvs93/}}.

\paragraph{Real-world datasets} We also included real-world multi-task datasets, encompassing both
classification and regression problems, carefully selected from established
multi-task studies~\citep{Zhao2018,Oneto2019,Wang2021,Wang2022} across diverse
domains, including image recognition~\citep{Cilia2020}, finance and
insurance~\citep{Moro2011}, healthcare~\citep{Gunduz2019,Srinivasan2024}, and
marketing~\citep{Argyriou2007}. These datasets present a variety of challenges,
including differences in the number of tasks, instances, and labels, as well as
varying degrees of task-relatedness. In each dataset, tasks are obtained by
grouping samples according to specific attribute values (e.g., demographic
categories or organizational units). Additional complexities arise from class
imbalances, skewed label distributions, and the presence of noise and outliers.
Table~\ref{tab:datasets} summarizes the real-world multi-task datasets along
with their corresponding references. Notably, the \textit{Avila} dataset
contains $12$ class labels, while the other classification datasets are binary.
For the \textit{Adult} dataset, tasks are defined by dividing the population
into gender and race categories, resulting in two separate datasets:
\textit{Adult-Gender} and \textit{Adult-Race}.

\begin{table}[ht]
      \centering
      \caption{Real-world multi-task datasets description.}
      \label{tab:datasets}
      \begin{tabularx}{\textwidth}{Xccc}
            \toprule
            \textbf{Name}                   & \textbf{Instances} & \textbf{Features} & \textbf{Tasks} \\
            \midrule
            \multicolumn{4}{l}{\textbf{Classification}}                                               \\
            Avila~\citep{Stefano2018}       & $20,867$           & $10$              & $48$           \\
            Adult~\citep{Becker1996}        & $48,842$           & $14$              & $7$            \\
            Bank Marketing~\citep{Moro2011} & $45,211$           & $16$              & $12$           \\
            Landmine~\citep{Yilmaz2018}     & $14,820$           & $9$               & $29$           \\
            \midrule
            \multicolumn{4}{l}{\textbf{Regression}}                                                   \\
            Abalone~\citep{Nash1994}        & $4,177$            & $8$               & $3$            \\
            Computer~\citep{Lenk1996}       & $3,800$            & $13$              & $190$          \\
            Parkinsons~\citep{Tsanas2009}   & $5,875$            & $19$              & $42$           \\
            SARCOS~\citep{Jawanpuria2015}   & $342,531$          & $21$              & $7$            \\
            School~\citep{bakker2003}       & $15,362$           & $10$              & $139$          \\
            \bottomrule
      \end{tabularx}
\end{table}

\subsection{Setup}\label{subsec:setup}
Experiments were implemented in \texttt{Python} using open-source libraries.
The \ac{RMB-CLE} implementation, synthetic data generators, and datasets are
publicly available\footnote{\href{https://github.com/GAA-UAM/RMB-CLE}
      {github.com/GAA-UAM/RMB-CLE}} to ensure reproducibility. All runs were executed
on a standard multi-core CPU server with 48~GB RAM.

Regarding the baseline methods, we benchmarked \ac{RMB-CLE} against several
state-of-the-art boosting methods, a natural comparison since our approach also
employs boosting as the local ensemble with supervised ensembling. This choice
is further motivated by the fact that ensemble methods based on decision trees
consistently achieve state-of-the-art performance on tabular data, particularly
in settings similar to those considered here, often outperforming deep learning
alternatives~\cite{Ravid2022}. Other approaches discussed in
Section~\ref{sec:related_work}, such as deep \ac{MTL} methods based on shared
representations and gradient-level optimization (e.g., multi-objective
formulations and gradient conflict resolution
techniques~\cite{Sener2018,Yu2020,Liu2021}), operate through fundamentally
different training dynamics and do not discover task clusters. Including them
as primary baselines would therefore conflate architectural and optimization
differences rather than isolate the effect of error-driven task clustering and
local ensembling. For completeness, however, we include a direct empirical
comparison with a representative deep multi-task learning model in
Subsection~\ref{subsec:synthetic}.

The baselines include both \ac{MT} and \ac{ST} variants. Dedicated \ac{MT}
models include \ac{MTGB}~\citep{Emami2023}, which consists of two training
blocks: the first uses common estimators to capture shared features across
tasks ($\mathcal{S}_{(1)}$), while the second includes task-specific estimators
to model task-specific patterns ($\mathcal{S}_{(2)}$), and
\ac{R-MTGB}~\citep{Emami2025}, which extends \ac{MTGB} by adding an
intermediate block ($\mathcal{S}_{(3)}$) to handle outliers using learnable
parameters $\sigma(\mathbf{\theta})$ that assign extreme weights, allowing the
model to emphasize or de-emphasize outliers relative to non-outliers as needed.
Heuristic \ac{MT} baselines included \ac{DP}, which merges all tasks into a
single dataset and trains one global model and \ac{TaF}, which augments the
input of the \ac{DP} approach by concatenating a one-hot encoding of the task
index so the model can learn task-specific adjustments. For \ac{ST} comparison,
\ac{GB}~\citep{Friedman2001} and \ac{LGBM}~\citep{Ke2017} were implemented.
Also, heuristic \ac{MT} variants were instantiated with \ac{GB} and \ac{LGBM}.
Finally, we employed both \ac{MTGB} and \ac{LGBM} as local ensembles in the
proposed \ac{RMB-CLE} framework, resulting in two variants: RMB-CLE-via-MTGB
and RMB-CLE-via-LGBM. To evaluate the ability of \ac{RMB-CLE} to identify
clusters of related tasks, we compared it with \emph{cluster-known} baselines
that were given the true task clusters. This comparison is limited to the
synthetic experiments and serves as an upper bound for assessing how closely
the proposed framework approximates optimal clustering performance.

Regarding the evaluation protocol, datasets have varying numbers of tasks and a
large number of experimental repetitions. For each of the 100 repetitions,
model performance on unseen data was first averaged across all tasks and
instances. We then computed the overall mean and \ac{Std Dev} across
repetitions. Each repetition corresponds to a different 80:20 train-test split
for real-world datasets or a different generated dataset for synthetic
experiments.

To ensure fair comparison across models, hyperparameters were selected using
within-training 5-fold cross-validation. Model configurations were chosen to
optimize negative \ac{RMSE} for regression and accuracy for classification. To
reduce variability from extensive tuning and maintain comparability, the grid
search was limited to the number of estimators (i.e., base learners). All other
hyperparameters were kept at their default values from the underlying
\texttt{scikit-learn}
implementation\footnote{\href{https://github.com/scikit-learn/scikit-learn}{github.com/scikit-learn}}~\citep{pedregosa2011scikit}.
For all ensemble-based models, decision tree regressors were used as base
learners with depth of one (i.e., decision stumps) to maintain consistency
across methods and limit the influence of individual tree complexity on overall
performance. The hyperparameter search space for all boosting-based baselines,
as well as for the proposed \ac{RMB-CLE} framework, includes the number of
predictors (block size ($K_{\mathcal{S}_{(.)}}$)) in each boosting block
($\mathcal{S}_{(.)}$), as defined in Subsection~\ref{subsec:preliminaries}. The
corresponding search space is summarized in
Table~\ref{tab:hyperparameter_search}. For \ac{R-MTGB}, three sequential
boosting blocks were tuned following the architecture proposed
in~\cite{Emami2025}, whereas \ac{MTGB} baseline was tuned using two blocks as
in~\cite{Emami2023}. Pooling-based baselines were tuned using a single global
boosting block trained on the pooled data from all tasks. Single-task baselines
(e.g., \ac{ST}-\ac{GB} and \ac{ST}-\ac{LGBM}) were tuned independently for each
task using a per-task boosting block. For \ac{RMB-CLE}, the number of base
learners in the local ensembles (Eq.~\eqref{eq:local-ensemble}) was selected
according to the underlying ensemble instantiation: a single shared learning
block across all tasks when using \ac{LGBM}, and two sequential blocks when
using \ac{MTGB}.
\begin{table}[t]
      \centering
      \caption{Hyperparameter search space for studied models.}
      \label{tab:hyperparameter_search}
      \begin{tabularx}{\textwidth}{l c X}
            \toprule
            \textbf{Method} & \textbf{Boosting block ($\mathcal{S}_{(.)}$)} & \textbf{Block size ($K_{\mathcal{S}_{(.)}}$)} \\
            \midrule
            Single-task approaches
                            & $\mathcal{S}_{(2)}$
                            & $\{20, 30, 50, 100\}$                                                                         \\
            \midrule
            Pooling-based approaches
                            & $\mathcal{S}_{(1)}$
                            & $\{20, 30, 50, 100\}$                                                                         \\
            \midrule
            MTGB
                            & $\mathcal{S}_{(1)}$
                            & $\{20, 30, 50\}$                                                                              \\
                            & $\mathcal{S}_{(2)}$
                            & $\{0, 20, 30, 50, 100\}$                                                                      \\
            \midrule
            R-MTGB
                            & $\mathcal{S}_{(1)}$
                            & $\{0, 20, 30, 50\}$                                                                           \\
                            & $\mathcal{S}_{(2)}$
                            & $\{0, 20, 30, 50, 100\}$                                                                      \\
                            & $\mathcal{S}_{(3)}$
                            & $\{20, 30, 50\}$                                                                              \\
            \midrule
            RMB-CLE-via-LGBM
                            & $\mathcal{S}_{(2)}$
                            & $\{20, 30, 50, 100\}$                                                                         \\
                            & $\mathcal{S}_{(4)}$
                            & $\{100\}$                                                                                     \\
            \midrule
            RMB-CLE-via-MTGB
                            & $\mathcal{S}_{(1)}$
                            & $\{20, 30, 50\}$                                                                              \\
                            & $\mathcal{S}_{(2)}$
                            & $\{0, 20, 30, 50, 100\}$                                                                      \\
                            & $\mathcal{S}_{(4)}$
                            & $\{100\}$                                                                                     \\
            \bottomrule
      \end{tabularx}
\end{table}

For statistical comparison, we perform a task-wise comparison of task-specific
models, with each model evaluated on the average performance of each task
across all repetitions. Accuracy is used for 96 classification tasks and
\ac{RMSE} for 381 regression tasks. Rankings in the Dems\v{a}r horizontal plots
(lower is better) are based on average ranks computed across tasks and compared
using the Nemenyi post-hoc test~\citep{demvsar2006statistical}. A statistically
significant difference between two models is indicated when they are not
connected by a solid black line, meaning their difference in average ranks
exceeds the corresponding \ac{CD}. We report average ranks separately for
synthetic and real-world benchmarks to distinguish controlled settings with
known latent task structure from heterogeneous real data.

\subsection{Synthetic results}~\label{subsec:synthetic}
Table~\ref{tab:synthetic_acc_rmse} summarizes the mean and \ac{Std Dev} of the
performance of the evaluated models on the multi-task synthetic classification
and regression datasets. The top scores are highlighted in \textbf{bold}. The
results show that RMB-CLE-via-LGBM consistently achieves the highest accuracy
($\approx 0.901$) and the lowest RMSE ($\approx 0.275$). Notably, the \ac{ST}
baseline, particularly ST-LGBM, outperforms most conventional \ac{MT} methods,
including \ac{MTGB}, \ac{R-MTGB}, and pooling-based strategies. This behavior
suggests that these \ac{MT} approaches are not robust to task heterogeneity,
and that their performance deteriorates in the presence of outlier or weakly
related tasks due to negative transfer. In comparison, \ac{RMB-CLE} mitigates
this issue by identifying and separating heterogeneous task groups prior to
knowledge sharing. While \ac{R-MTGB} partially improves robustness by
distinguishing outlier from non-outlier tasks, it is limited to a binary
partition.

\begin{table}[!ht]
      \centering
      \caption{Synthetic multi-task results (mean ± Std Dev).
            Accuracy (classification) and RMSE (regression); Best results in bold.}
      \label{tab:synthetic_acc_rmse}
      \begin{tabularx}{\textwidth}{Xcc}
            \toprule
            \textbf{Model}     & \textbf{Accuracy}          & \textbf{RMSE}              \\
            \midrule
            RMB-CLE-via-LGBM   & \textbf{0.900 $\pm$ 0.015} & \textbf{0.278 $\pm$ 0.017} \\
            Cluster-Known-LGBM & \textbf{0.900 $\pm$ 0.015} & \textbf{0.278 $\pm$ 0.017} \\
            ST-LGBM            & 0.789 $\pm$ 0.026          & 0.450 $\pm$ 0.032          \\
            RMB-CLE-via-MTGB   & 0.767 $\pm$ 0.029          & 0.550 $\pm$ 0.039          \\
            Cluster-Known-MTGB & 0.767 $\pm$ 0.029          & 0.550 $\pm$ 0.039          \\
            ST-GB              & 0.759 $\pm$ 0.028          & 0.549 $\pm$ 0.038          \\
            MTGB               & 0.698 $\pm$ 0.079          & 0.783 $\pm$ 0.143          \\
            R-MTGB             & 0.690 $\pm$ 0.074          & 0.777 $\pm$ 0.102          \\
            DP-LGBM            & 0.616 $\pm$ 0.039          & 0.840 $\pm$ 0.113          \\
            TaF-LGBM           & 0.616 $\pm$ 0.038          & 0.791 $\pm$ 0.089          \\
            DP-GB              & 0.611 $\pm$ 0.038          & 0.841 $\pm$ 0.113          \\
            TaF-GB             & 0.611 $\pm$ 0.038          & 0.800 $\pm$ 0.090          \\
            \bottomrule
      \end{tabularx}
\end{table}

Moreover, to further substantiate the trends observed in
Table~\ref{tab:synthetic_acc_rmse}, we report complementary evaluation metrics
in Table~\ref{tab:synthetic_recall_mae}. In particular, this table presents
macro-averaged classification recall (equal weight across classes) and the
regression \ac{MAE}. The best results for each dataset are highlighted in
\textbf{bold}.

The results indicate that RMB-CLE-via-LGBM achieves the highest macro recall
($\approx 0.901$) and the lowest MAE ($\approx 0.215$), confirming the
consistent superiority of the proposed framework when leveraging LGBM. The
close correspondence between recall and accuracy values, as observed in
Table~\ref{tab:synthetic_acc_rmse}, demonstrates the robustness of the model,
indicating that improvements in sensitivity are not achieved at the expense of
overall predictive performance. On the other hand, baseline methods exhibit
lower recall and significantly higher MAE, emphasizing the stability and
effectiveness of the proposed clustering-based multi-task ensembling approach.

\begin{table}[ht]
      \centering
      \caption{Synthetic multi-task results (mean ± Std Dev).
            Recall (classification) and MAE (regression); Best results in bold.}
      \label{tab:synthetic_recall_mae}
      \small
      \setlength{\tabcolsep}{6pt}
      \begin{tabularx}{\textwidth}{Xcc}
            \toprule
            \textbf{Model}     & \textbf{Recall}            & \textbf{MAE}               \\
            \midrule
            RMB-CLE-via-LGBM   & \textbf{0.901 $\pm$ 0.016} & \textbf{0.218 $\pm$ 0.013} \\
            Cluster-Known-LGBM & \textbf{0.901 $\pm$ 0.016} & \textbf{0.218 $\pm$ 0.013} \\
            ST-LGBM            & 0.789 $\pm$ 0.026          & 0.351 $\pm$ 0.025          \\
            ST-GB              & 0.759 $\pm$ 0.028          & 0.434 $\pm$ 0.030          \\
            RMB-CLE-via-MTGB   & 0.767 $\pm$ 0.029          & 0.435 $\pm$ 0.031          \\
            Cluster-Known-MTGB & 0.767 $\pm$ 0.029          & 0.435 $\pm$ 0.031          \\
            R-MTGB             & 0.690 $\pm$ 0.074          & 0.624 $\pm$ 0.086          \\
            MTGB               & 0.698 $\pm$ 0.079          & 0.630 $\pm$ 0.121          \\
            TaF-LGBM           & 0.616 $\pm$ 0.038          & 0.636 $\pm$ 0.076          \\
            TaF-GB             & 0.611 $\pm$ 0.038          & 0.643 $\pm$ 0.076          \\
            DP-LGBM            & 0.616 $\pm$ 0.039          & 0.676 $\pm$ 0.093          \\
            DP-GB              & 0.611 $\pm$ 0.038          & 0.677 $\pm$ 0.093          \\
            \bottomrule
      \end{tabularx}
\end{table}

Furthermore, in Tables~\ref{tab:synthetic_acc_rmse}
and~\ref{tab:synthetic_recall_mae}, the oracle \emph{Cluster-Known} baselines,
which are given direct access to the ground-truth task clusters and therefore
bypass the clustering step, achieve identical performance to the corresponding
\ac{RMB-CLE} variants. This result indicates that the proposed framework
effectively recovers the true latent task structure from cross-task error
information, reaching the same performance as an idealized method with perfect
cluster knowledge.

However, conventional \ac{MTL} approaches exhibit poor performance, often
failing to outperform \ac{ST} baselines. As shown in
Tables~\ref{tab:synthetic_acc_rmse} and~\ref{tab:synthetic_recall_mae},
standard multi-task boosting methods such as \ac{MTGB} and \ac{R-MTGB} suffer
performance degradation when the task landscape comprises multiple
heterogeneous clusters, despite modeling shared and task-specific components.
Single-task models (e.g., \ac{ST}-\ac{LGBM}) consistently outperform these
\ac{MT} baselines, indicating that naïve or insufficiently structured
information sharing can be more harmful than no sharing at all.

\begin{figure}[ht]
      \centering
      \includegraphics[width=1.0\linewidth]{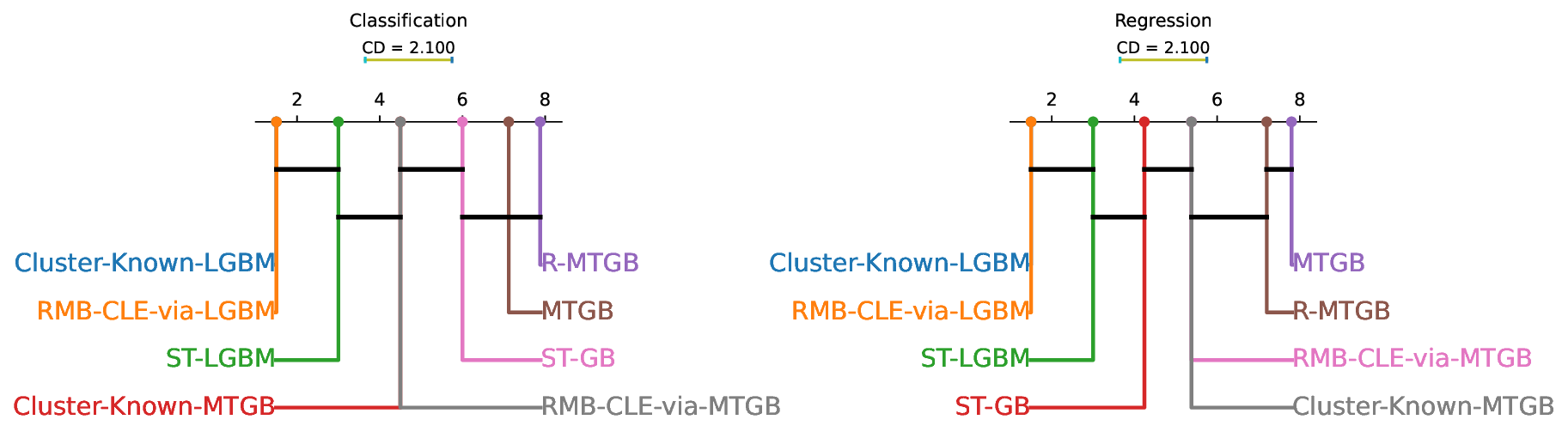}
      \caption{Synthetic task-wise Dem\v{s}ar plots ($p=0.05$) comparing multi-task models; solid
            lines indicate no significant differences (Nemenyi).}
      \label{fig:Figure_2}
\end{figure}

To further analyze performance at the task level, the rankings of the synthetic
tasks are shown in Figure~\ref{fig:Figure_2}: classification tasks in the left
subplot and regression tasks in the right subplot. The proposed framework
achieved performance identical to the upper bound, represented by
\emph{Cluster-Known}, for both \ac{MTGB} and \ac{LGBM} as local ensembles.
Moreover, the proposed framework with \ac{LGBM} obtained the best ranking,
followed by \ac{ST}-\ac{LGBM} and the proposed approach with \ac{MTGB}, across
all problems (all subplots). We observe that the proposed framework with
\ac{LGBM} is not statistically different from \ac{ST}-\ac{LGBM} in both
classification and regression problems. However, it shows statistically
significant differences compared to all other models. On the other hand, the
conventional boosting-based multi-task models (\ac{R-MTGB} and \ac{MTGB})
consistently achieved the lowest rankings across all tasks (all subplots).

To complement the boosting-based comparisons presented in the main experiments,
we include an additional baseline based on deep \ac{MTL} with shared
representations. This experiment is intended to assess how a representative
neural multi-task model compares to the proposed framework (under the same
synthetic dataset generated and used to train and evaluate the studied models),
while acknowledging the fundamentally different modeling assumptions.

We adopt a shared neural architecture implemented using \texttt{scikit-learn},
consisting of a \ac{MLP} that learns a common feature representation across all
tasks (Deep-MTL). The shared network comprises three fully connected hidden
layers, each with 100 units and ReLU activation functions. This shared
representation is followed by task-specific output heads, with one head per
task. For regression problems, each task head is implemented as a ridge
regression model, while for classification problems, task heads are implemented
as logistic regression classifiers.

The shared \ac{MLP} is trained on pooled data from all tasks without using task
identifiers as input features, thereby enforcing full parameter sharing in the
representation layers. Training is performed using the \textit{Adam} optimizer
with a learning rate of $10^{-3}$, a maximum of 100 iterations, and early
stopping enabled. The $\ell_2$ regularization parameter is fixed to $10^{-4}$.
After training, the learned hidden representation (defined as the output of the
final hidden layer) is extracted and used as input to train task-specific heads
independently for each task. For these heads, ridge regression with
regularization parameter $\alpha = 1.0$ is used in regression settings, while
logistic regression with inverse regularization strength equal to $1.0$ is used
for classification. All experiments use the same random seeds, data splits, and
evaluation protocol as in previous experiments.

Tables~\ref{tab:synthetic_acc_rmse_deepmtl}
and~\ref{tab:synthetic_recall_mae_deepmtl} report the performance of a
representative Deep-\ac{MTL} model based on a shared neural representation.
Compared to the boosting-based baselines reported in
Tables~\ref{tab:synthetic_acc_rmse} and~\ref{tab:synthetic_recall_mae},
Deep-MTL model achieves moderate performance in both classification and
regression. While the model benefits from parameter sharing across tasks, its
accuracy and recall remain below those of \ac{RMB-CLE}, and its regression
errors (\ac{RMSE} and \ac{MAE}) are notably higher.

\begin{table}[!ht]
      \centering
      \caption{Synthetic multi-task performance of the deep MTL baseline (mean ± Std Dev).
            Accuracy (classification) and RMSE (regression).
            Results are obtained using the same synthetic data, splits, and evaluation protocol as
            in Tables~\ref{tab:synthetic_acc_rmse} and~\ref{tab:synthetic_recall_mae}.}
      \label{tab:synthetic_acc_rmse_deepmtl}
      \begin{tabularx}{\textwidth}{Xcc}
            \toprule
            \textbf{Model} & \textbf{Accuracy} & \textbf{RMSE}     \\
            \midrule
            Deep-MTL       & $0.848 \pm 0.015$ & $0.313 \pm 0.022$ \\
            \bottomrule
      \end{tabularx}
\end{table}

\begin{table}[!ht]
      \centering
      \caption{Synthetic multi-task performance of the deep MTL baseline (mean ± Std Dev).
            Recall (classification) and MAE (regression).
            Results are obtained using the same synthetic data, splits, and evaluation protocol as in Tables~\ref{tab:synthetic_acc_rmse} and~\ref{tab:synthetic_recall_mae}.}
      \label{tab:synthetic_recall_mae_deepmtl}
      \begin{tabularx}{\textwidth}{Xcc}
            \toprule
            \textbf{Model} & \textbf{Recall}   & \textbf{MAE}      \\
            \midrule
            Deep-MTL       & $0.848 \pm 0.015$ & $0.244 \pm 0.017$ \\
            \bottomrule
      \end{tabularx}
\end{table}

To further investigate the robustness of our method, we conducted an additional
ablation study (reported in~\ref{appendix:alt_linkage}) to evaluate the
sensitivity of the clustering stage to the choice of linkage criterion. Average
and complete linkage differ in how inter-cluster distances are defined during
hierarchical agglomerative clustering: average linkage relies on the mean
pairwise distance between tasks across clusters, whereas complete linkage
considers the maximum such distance, resulting in more compact
clusters~\cite{Rokach2005, Murtagh2017}. Despite these differences, both
strategies yield identical performance (see~\ref{appendix:alt_linkage}),
indicating that the proposed error-based similarity representation produces a
well-structured and stable task geometry. Consequently, the recovered task
groups are robust to the specific rule used for cluster merging.

Moreover, we measured the elapsed runtime of the studied models under three
different scenarios on the generated synthetic datasets, with detailed results
reported in~\ref{apendix:time_tables}. The experiments indicate that
pooling-based models with \ac{LGBM} are the fastest, whereas multi-task
boosting methods (\ac{MTGB}, \ac{R-MTGB}) have substantially higher training
times due to their sequential multi-block architectures. Notably,
RMB-CLE-via-LGBM matches the \emph{Cluster-Known} baseline in total training
time, remaining considerably faster than \ac{MTGB} and \ac{R-MTGB}, while
RMB-CLE-via-MTGB exhibits the higher runtime characteristic of multi-task
boosting.

\subsubsection{Cluster stability analysis}~\label{subsubsec:cluster_stability}
\begin{figure}[htb]
      \centering
      \includegraphics[width=0.95\linewidth]{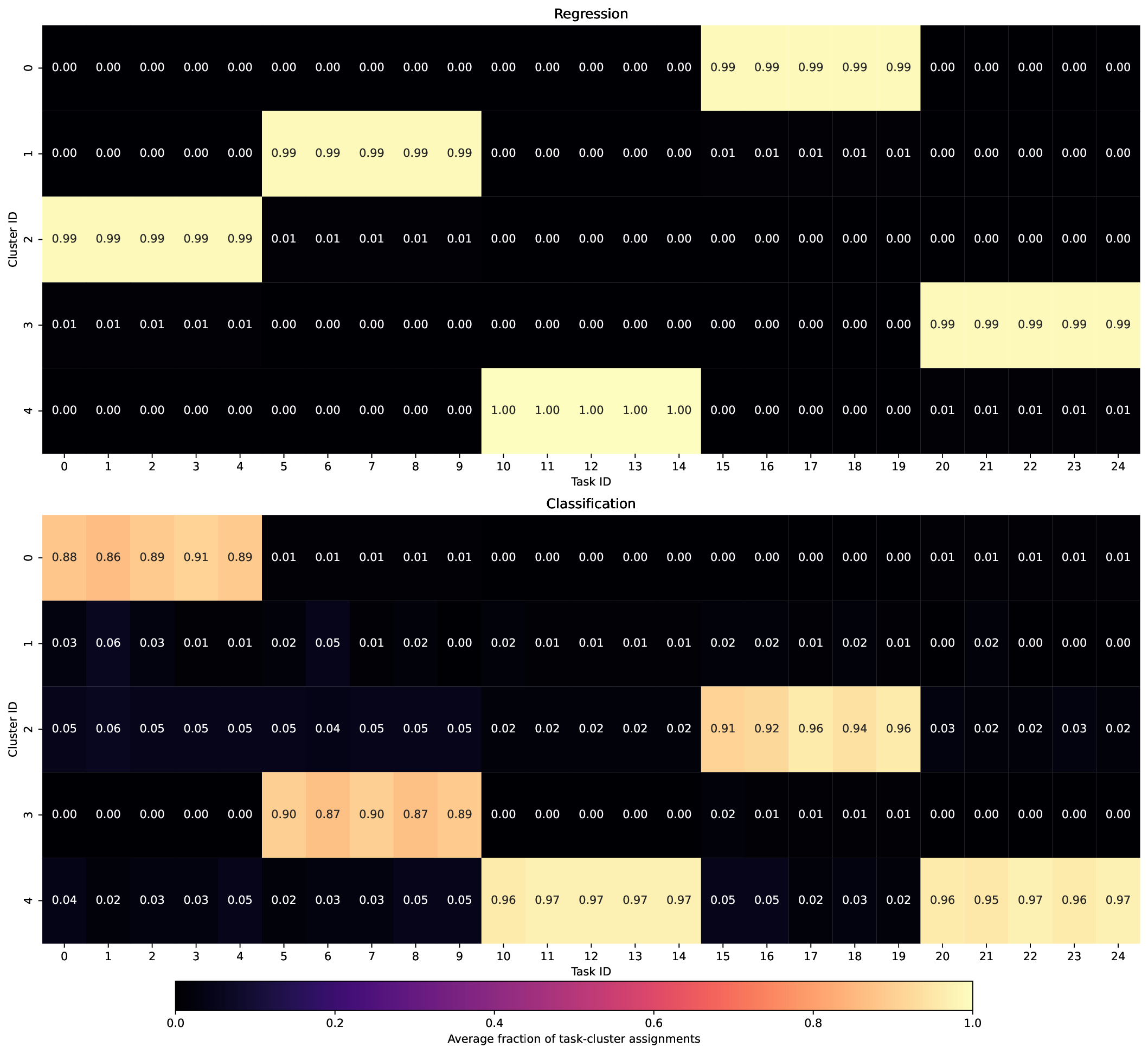}
      \caption{Cluster assignment stability over 100 runs.
            Each panel shows the fraction of times a task is assigned to each inferred cluster.
            Columns denote tasks and rows denote clusters; brighter colors indicate more stable
            assignments. Ground-truth clusters are reported in Table~\ref{tab:true_clusters}.}
      \label{fig:Figure_3}
\end{figure}

To assess the stability of the inferred task clusters,
Figure~\ref{fig:Figure_3} reports the frequency with which each
task is assigned to each cluster over 100 independent runs by
the proposed framework. For both regression (top panel) and
classification (bottom panel). In each heatmap (panel), columns represent individual tasks and
rows correspond to cluster indices. Each cell shows the fraction of runs in
which a given task was assigned to a particular cluster, with higher color
intensity indicating more frequent assignments.
Figure~\ref{fig:Figure_3} reveals that
tasks that belong to the same ground-truth cluster (Table~\ref{tab:true_clusters})
are consistently mapped to the same inferred cluster,
with assignment frequencies close to one, while assignments to
other clusters remain near zero.
This pattern indicates that the clustering procedure
is not sensitive to random initialization or data splits
and reliably recovers the latent task structure embedded
in the synthetic data. Minor dispersion observed in
the classification setting reflects the higher
stochasticity induced by thresholded labels, yet
the dominant cluster memberships remain stable across runs.

\begin{table}[t]
      \centering
      \caption{Ground-truth task-to-cluster assignments
            in the synthetic data.}~\label{tab:true_clusters}
      \begin{tabular}{cl}
            \toprule
            \textbf{Cluster ID} & \textbf{Task indices} \\
            \midrule
            0                   & $[0,1,2,3,4]$         \\
            1                   & $[5,6,7,8,9]$         \\
            2                   & $[10,11,12,13,14]$    \\
            3                   & $[15,16,17,18,19]$    \\
            4                   & $[20,21,22,23,24]$    \\
            \bottomrule
      \end{tabular}
\end{table}

Moreover, Table~\ref{tab:synthetic_acc_rmse} quantitatively confirms this
observation, showing that the proposed \ac{RMB-CLE} framework achieves
performance identical to the oracle \emph{Cluster-Known} baseline, which has
access to the true task-to-cluster assignments. The coincidence of predictive
performance implies that the clusters inferred by \ac{RMB-CLE} are effectively
equivalent to the ground-truth clusters for the purpose of learning.

\begin{figure}[htb]
      \centering
      \includegraphics[width=1.0\linewidth]{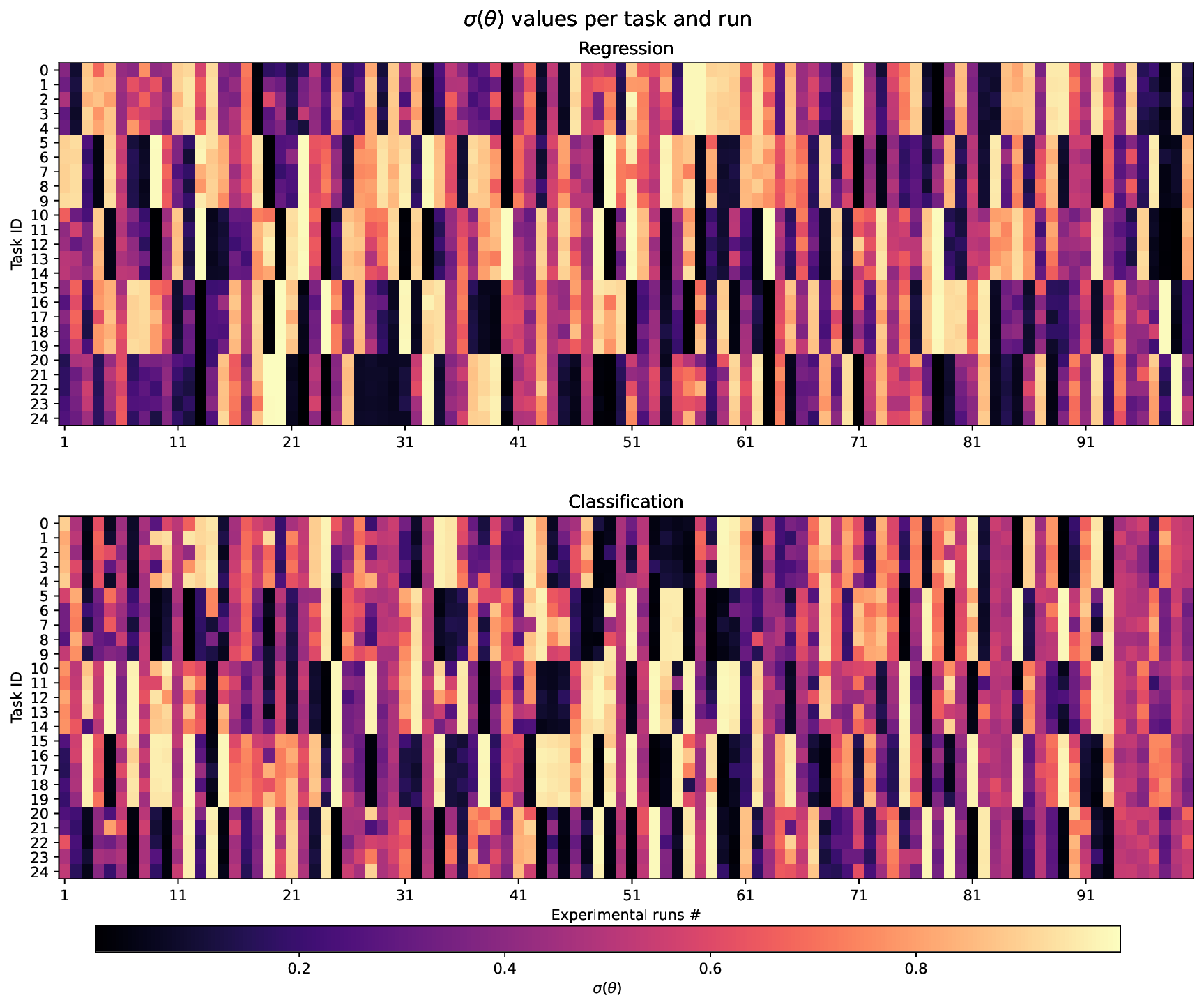}
      \caption{Stability of sigmoid-based task weighting in R-MTGB over 100 runs.
            Each panel shows the learned task ($i$) weights $\sigma(\theta_i)$ across runs.
            Columns denote runs and rows denote tasks; color intensity reflects the magnitude
            of $\sigma(\theta_i)$, indicating how strongly tasks are weighted.}~\label{fig:Figure_4}
\end{figure}

To contextualize the clustering behavior of the \ac{R-MTGB} model, we analyze
the task-wise behavior of the sigmoid-based weighting mechanism
($\sigma(\theta_i)$) employed in \ac{R-MTGB} across multiple runs. In
\ac{R-MTGB}, $\theta_i$ is a learnable, task-specific scalar parameter that
controls the relative contribution of each task to the shared boosting stage by
softly gating tasks as either inliers or outliers. Through the sigmoid
transformation,
\[
      \sigma(\theta_i) = \frac{1}{1 + \exp(-\theta_i)},
\]
each task is assigned a continuous weight in the interval
$(0,1)$~\cite{Emami2025}. Values of $\sigma(\boldsymbol{\theta})$ close to $1$
indicate tasks that are coupled to the shared representation (inliers), while
values close to $0$ correspond to tasks that are effectively down-weighted and
treated as outliers. Depending on the sign convention of $\theta_i$, this
interpretation may be reversed, but the underlying behavior remains unchanged.
If the underlying task structure aligns with the assumptions of \ac{R-MTGB},
the learned $\boldsymbol{\theta}$ values are expected to be stable across runs
and to concentrate near the extremes. However, when the true task landscape
contains multiple coherent clusters, the use of a single scalar parameter
$\theta_i$ per task becomes too limited to capture this structure. In this
case, $\boldsymbol{\theta}$ can only encode a one-dimensional gating decision
and is therefore unable to represent richer multi-cluster structure.
Consequently, the sigmoid weighting mechanism ($\sigma(\boldsymbol{\theta})$)
is forced to collapse several distinct task groups into an soft binary
partition. As a result, and as illustrated in Figure~\ref{fig:Figure_4}, the
learned $\boldsymbol{\theta}$ values exhibit substantial variability across
runs and tasks, failing to produce a stable or interpretable task separation.

Figure~\ref{fig:Figure_4} highlights this limitation by visualizing the learned
$\sigma(\boldsymbol{\theta})$ values across tasks and runs for both regression
(top panel) and classification (bottom panel). In this figure, columns
correspond to different runs and rows to tasks, with color intensity
representing the magnitude of $\sigma(\theta_i)$ for each task. Rather than
revealing a consistent pattern that reflects the underlying multi-cluster
structure of the synthetic data, Figure~\ref{fig:Figure_4} shows highly
heterogeneous and unstable weight assignments in the heatmaps. Tasks that
belong to the same ground-truth cluster often receive different weights across
runs, while tasks from distinct clusters frequently overlap in their
$\sigma(\boldsymbol{\theta})$ values. This lack of coherence indicates that the
sigmoid gating mechanism does not converge to a meaningful partition when more
than two task groups are present. Instead, the optimization oscillates between
competing binary explanations of the task landscape, leading to run-dependent
and task-agnostic weight configurations.

\subsection{Real-world results}~\label{subsec:realworld}
Regarding the real-world datasets,
Tables~\ref{tab:realworld_clf_performance}
and~\ref{tab:realworld_clf_performance_recall} report classification
performance in terms of accuracy and macro-averaged recall, respectively, while
Tables~\ref{tab:realworld_reg_performance} and~\ref{tab:realworld_reg_performance_mae}
summarize regression results using \ac{RMSE}
and \ac{MAE}. All results are reported as the mean and \ac{Std Dev},
where performance is first averaged across all tasks for each dataset
and then averaged over 100 independent runs.
The best-performing method for each dataset is
highlighted in \textbf{bold}.

Consistent with the synthetic experiments, the proposed framework utilizing
\ac{LGBM} achieves superior results across all classification datasets
(Tables~\ref{tab:realworld_clf_performance}
and~\ref{tab:realworld_clf_performance_recall}). The low \ac{Std Dev} values
indicate that this improvement is not at the expense of stability. In addition,
we computed the F1-score (reported in~\ref{apendix:F1_score}) to assess model
performance under class imbalance, and the results show that the proposed
framework consistently achieves superior precision-recall balance compared to
the studied baselines.

\begin{table*}[!ht]
      \centering
      \caption{Real-world classification accuracy (mean ± Std Dev).
            Best results in bold.}
      \label{tab:realworld_clf_performance}
      \resizebox{\textwidth}{!}{%
            \begin{tabular}{lccccc}
                  \toprule
                  \textbf{Model}   & \textbf{Adult-Gender}               & \textbf{Adult-Race}                 & \textbf{Avila}                      & \textbf{Bank}                       & \textbf{Landmine}                   \\
                  \midrule
                  DP-GB            & 0.837 $\pm$ 0.005                   & 0.837 $\pm$ 0.005                   & 0.494 $\pm$ 0.010                   & 0.889 $\pm$ 0.003                   & 0.939 $\pm$ 0.003                   \\
                  DP-LGBM          & 0.854 $\pm$ 0.004                   & 0.854 $\pm$ 0.004                   & 0.536 $\pm$ 0.021                   & 0.898 $\pm$ 0.003                   & 0.939 $\pm$ 0.003                   \\
                  RMB-CLE-via-LGBM & \textbf{0.872} $\pm$ \textbf{0.003} & \textbf{0.873} $\pm$ \textbf{0.003} & \textbf{0.875} $\pm$ \textbf{0.149} & \textbf{0.906} $\pm$ \textbf{0.003} & \textbf{0.946} $\pm$ \textbf{0.004} \\
                  RMB-CLE-via-MTGB & 0.848 $\pm$ 0.004                   & 0.837 $\pm$ 0.007                   & 0.604 $\pm$ 0.055                   & 0.894 $\pm$ 0.003                   & 0.943 $\pm$ 0.004                   \\
                  MTGB             & 0.848 $\pm$ 0.004                   & 0.845 $\pm$ 0.004                   & 0.614 $\pm$ 0.050                   & 0.893 $\pm$ 0.003                   & 0.943 $\pm$ 0.004                   \\
                  R-MTGB           & 0.849 $\pm$ 0.004                   & 0.849 $\pm$ 0.004                   & 0.619 $\pm$ 0.047                   & 0.895 $\pm$ 0.003                   & 0.943 $\pm$ 0.004                   \\
                  ST-GB            & 0.841 $\pm$ 0.004                   & 0.838 $\pm$ 0.004                   & 0.610 $\pm$ 0.060                   & 0.892 $\pm$ 0.003                   & 0.942 $\pm$ 0.003                   \\
                  ST-LGBM          & 0.855 $\pm$ 0.003                   & 0.852 $\pm$ 0.004                   & 0.690 $\pm$ 0.092                   & 0.897 $\pm$ 0.003                   & 0.940 $\pm$ 0.003                   \\
                  TaF-GB           & 0.837 $\pm$ 0.005                   & 0.837 $\pm$ 0.005                   & 0.497 $\pm$ 0.009                   & 0.889 $\pm$ 0.003                   & 0.939 $\pm$ 0.003                   \\
                  TaF-LGBM         & 0.854 $\pm$ 0.004                   & 0.854 $\pm$ 0.004                   & 0.552 $\pm$ 0.114                   & 0.898 $\pm$ 0.003                   & 0.939 $\pm$ 0.003                   \\
                  \bottomrule
            \end{tabular}
      }
\end{table*}

\begin{table}[ht]
      \centering
      \small
      \caption{Real-world classification recall (mean ± Std Dev).
            Best results in bold.}
      \label{tab:realworld_clf_performance_recall}
      \resizebox{1.0\textwidth}{!}{%
            \begin{tabular}{lcccccc}
                  \toprule
                  \textbf{Model}   & \textbf{Adult-Gender}      & \textbf{Adult-Race}        & \textbf{Avila}             & \textbf{Bank}              & \textbf{Landmine}          \\
                  \midrule
                  DP-GB            & 0.684 $\pm$ 0.010          & 0.684 $\pm$ 0.010          & 0.166 $\pm$ 0.008          & 0.532 $\pm$ 0.004          & 0.500 $\pm$ 0.000          \\
                  DP-LGBM          & 0.739 $\pm$ 0.005          & 0.739 $\pm$ 0.005          & 0.263 $\pm$ 0.025          & 0.610 $\pm$ 0.006          & 0.503 $\pm$ 0.003          \\
                  RMB-CLE-via-LGBM & \textbf{0.796 $\pm$ 0.005} & \textbf{0.797 $\pm$ 0.005} & \textbf{0.774 $\pm$ 0.182} & \textbf{0.708 $\pm$ 0.010} & \textbf{0.591 $\pm$ 0.014} \\
                  RMB-CLE-via-MTGB & 0.720 $\pm$ 0.005          & 0.684 $\pm$ 0.018          & 0.446 $\pm$ 0.083          & 0.587 $\pm$ 0.011          & 0.552 $\pm$ 0.010          \\
                  MTGB             & 0.720 $\pm$ 0.005          & 0.716 $\pm$ 0.008          & 0.438 $\pm$ 0.085          & 0.576 $\pm$ 0.010          & 0.549 $\pm$ 0.010          \\
                  R-MTGB           & 0.726 $\pm$ 0.006          & 0.725 $\pm$ 0.005          & 0.436 $\pm$ 0.083          & 0.590 $\pm$ 0.009          & 0.548 $\pm$ 0.011          \\
                  ST-GB            & 0.705 $\pm$ 0.006          & 0.693 $\pm$ 0.008          & 0.454 $\pm$ 0.088          & 0.555 $\pm$ 0.005          & 0.552 $\pm$ 0.010          \\
                  ST-LGBM          & 0.744 $\pm$ 0.005          & 0.737 $\pm$ 0.005          & 0.589 $\pm$ 0.122          & 0.617 $\pm$ 0.007          & 0.524 $\pm$ 0.007          \\
                  TaF-GB           & 0.684 $\pm$ 0.010          & 0.684 $\pm$ 0.010          & 0.169 $\pm$ 0.007          & 0.532 $\pm$ 0.004          & 0.500 $\pm$ 0.000          \\
                  TaF-LGBM         & 0.739 $\pm$ 0.005          & 0.739 $\pm$ 0.005          & 0.347 $\pm$ 0.135          & 0.610 $\pm$ 0.006          & 0.502 $\pm$ 0.003          \\
                  \bottomrule
            \end{tabular}
      }
\end{table}

For regression datasets, Tables~\ref{tab:realworld_reg_performance}
and~\ref{tab:realworld_reg_performance_mae} show that \ac{MT} approaches,
particularly RMB-CLE-via-LGBM, generally achieve the lowest \ac{RMSE},
confirming the advantage of task clustering. An exception is the
\textit{Parkinson} dataset, where \ac{ST}-\ac{GB} attains the best result, with
a marginal. Notably, \ac{DP} approaches perform poorly in datasets with many
tasks (e.g., \textit{SARCOS} and \textit{Parkinson}).

\begin{table*}[htb]
      \centering
      \caption{Real-world regression RMSE (mean ± Std Dev).
            Best results in bold.}
      \label{tab:realworld_reg_performance}
      \resizebox{\textwidth}{!}{%
            \begin{tabular}{lccccc}
                  \toprule
                  \textbf{Model}   & \textbf{Abalone}                  & \textbf{Computer}                 & \textbf{Parkinson}                & \textbf{SARCOS}                   & \textbf{School}                    \\
                  \midrule
                  DP-GB            & 2.397$\pm$0.093                   & 2.466$\pm$0.048                   & 8.859$\pm$0.137                   & 18.397$\pm$0.067                  & 10.423$\pm$0.118                   \\
                  DP-LGBM          & 2.403$\pm$0.093                   & 2.466$\pm$0.048                   & 8.859$\pm$0.136                   & 18.395$\pm$0.067                  & 10.424$\pm$0.118                   \\
                  RMB-CLE-via-LGBM & \textbf{2.169}$\pm$\textbf{0.085} & \textbf{2.420}$\pm$\textbf{0.087} & 0.349$\pm$0.035                   & \textbf{2.507}$\pm$\textbf{0.020} & \textbf{10.056}$\pm$\textbf{0.116} \\
                  RMB-CLE-via-MTGB & 2.465$\pm$0.081                   & 2.754$\pm$0.333                   & 1.277$\pm$0.085                   & 10.776$\pm$0.049                  & 10.509$\pm$0.124                   \\
                  MTGB             & 2.289$\pm$0.087                   & 2.486$\pm$0.047                   & 0.336$\pm$0.024                   & 4.808$\pm$0.034                   & 10.154$\pm$0.122                   \\
                  R-MTGB           & 2.266$\pm$0.086                   & 2.463$\pm$0.076                   & 0.289$\pm$0.037                   & 4.698$\pm$0.066                   & 10.133$\pm$0.119                   \\
                  ST-GB            & 2.346$\pm$0.089                   & 2.760$\pm$0.352                   & \textbf{0.268}$\pm$\textbf{0.027} & 4.919$\pm$0.034                   & 10.295$\pm$0.137                   \\
                  ST-LGBM          & 2.345$\pm$0.089                   & 2.948$\pm$0.214                   & 0.557$\pm$0.035                   & 4.926$\pm$0.034                   & 10.669$\pm$0.131                   \\
                  TaF-GB           & 2.383$\pm$0.093                   & 2.467$\pm$0.067                   & 6.559$\pm$0.087                   & 11.266$\pm$0.060                  & 10.415$\pm$0.117                   \\
                  TaF-LGBM         & 2.388$\pm$0.093                   & 2.469$\pm$0.048                   & 6.558$\pm$0.087                   & 11.266$\pm$0.060                  & 10.416$\pm$0.118                   \\
                  \bottomrule
            \end{tabular}
      }
\end{table*}

\begin{table}[htb]
      \centering
      \small
      \caption{Real-world regression MAE (mean ± Std Dev).
            Best results in bold.}
      \label{tab:realworld_reg_performance_mae}
      \resizebox{1.0\textwidth}{!}{%
            \begin{tabular}{lcccccc}
                  \toprule
                  \textbf{Model}   & \textbf{Abalone}           & \textbf{Computer}          & \textbf{Parkinson}         & \textbf{SARCOS}            & \textbf{School}            \\
                  \midrule
                  DP-GB            & 1.732 $\pm$ 0.050          & 2.025 $\pm$ 0.043          & 7.340 $\pm$ 0.118          & 12.650 $\pm$ 0.047         & 8.270 $\pm$ 0.097          \\
                  DP-LGBM          & 1.737 $\pm$ 0.051          & 2.025 $\pm$ 0.043          & 7.339 $\pm$ 0.117          & 12.648 $\pm$ 0.047         & 8.270 $\pm$ 0.098          \\
                  RMB-CLE-via-LGBM & \textbf{1.524 $\pm$ 0.046} & \textbf{1.972 $\pm$ 0.071} & 0.203 $\pm$ 0.014          & \textbf{1.473 $\pm$ 0.012} & \textbf{7.938 $\pm$ 0.093} \\
                  RMB-CLE-via-MTGB & 1.882 $\pm$ 0.070          & 2.200 $\pm$ 0.298          & 0.850 $\pm$ 0.059          & 7.159 $\pm$ 0.028          & 8.351 $\pm$ 0.109          \\
                  MTGB             & 1.624 $\pm$ 0.047          & 2.054 $\pm$ 0.044          & 0.186 $\pm$ 0.015          & 2.778 $\pm$ 0.016          & 8.031 $\pm$ 0.095          \\
                  R-MTGB           & 1.607 $\pm$ 0.046          & 2.020 $\pm$ 0.068          & 0.137 $\pm$ 0.029          & 2.734 $\pm$ 0.033          & 8.006 $\pm$ 0.094          \\
                  ST-GB            & 1.664 $\pm$ 0.048          & 2.197 $\pm$ 0.317          & \textbf{0.110 $\pm$ 0.008} & 2.778 $\pm$ 0.015          & 8.146 $\pm$ 0.108          \\
                  ST-LGBM          & 1.664 $\pm$ 0.048          & 2.432 $\pm$ 0.194          & 0.301 $\pm$ 0.015          & 2.780 $\pm$ 0.016          & 8.460 $\pm$ 0.112          \\
                  TaF-GB           & 1.711 $\pm$ 0.050          & 2.043 $\pm$ 0.056          & 5.713 $\pm$ 0.088          & 7.132 $\pm$ 0.031          & 8.270 $\pm$ 0.098          \\
                  TaF-LGBM         & 1.715 $\pm$ 0.050          & 2.028 $\pm$ 0.043          & 5.712 $\pm$ 0.088          & 7.130 $\pm$ 0.030          & 8.269 $\pm$ 0.098          \\
                  \bottomrule
            \end{tabular}
      }
\end{table}

\begin{figure}[ht]
      \centering
      \includegraphics[width=1.0\linewidth]{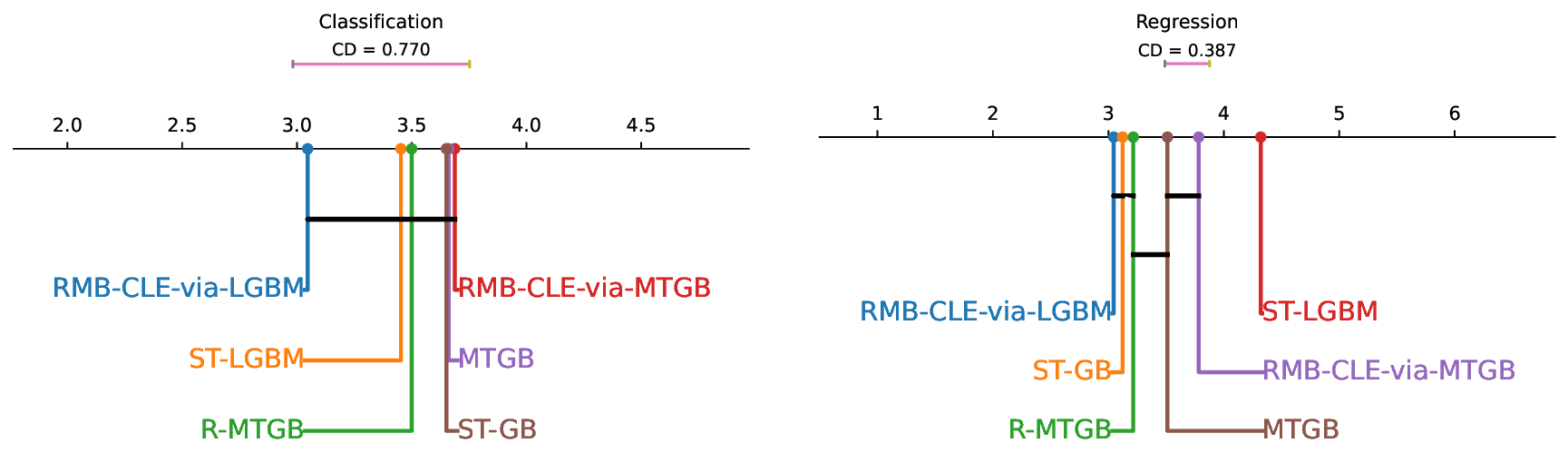}
      \caption{Real-world task-wise Dem\v{s}ar plots ($p=0.05$) comparing multi-task models; solid
            lines indicate no significant differences (Nemenyi).}~\label{fig:Figure_6}
\end{figure}

To further assess performance at the task level on real-world datasets, we
follow the same task-wise statistical comparison protocol used in the synthetic
experiments (Subsection~\ref{subsec:synthetic}) and analyze the relative
ranking of methods across tasks. Figure~\ref{fig:Figure_6} reports the average
ranks of task-wise methods computed over individual tasks, where lower ranks
indicate better performance. Horizontal bars connect methods whose differences
in average rank are not statistically significant at the $5\%$ level, as
determined by the \ac{CD}. The results show that, consistent with the synthetic
results, the proposed approach achieves the best average rank across both
classification (left subplot) and regression tasks (right subplot), and is
statistically separated from most competing baselines in regression setting.
Among the remaining methods, \ac{ST}-\ac{LGBM} attains the next best ranking
for classification problem, while \ac{ST}-\ac{GB} performs best among the
baselines for regression problem, though both remain inferior to the proposed
framework in terms of average rank.

\section{Conclusions}~\label{sec:conclusions}
This work introduced \acf{RMB-CLE}, a robust multi-task boosting framework that
reframes task relatedness as a question of cross-task generalization rather
than in-domain performance or heuristic similarity. By grounding task
similarity in error-driven transfer behavior and combining it with adaptive
hierarchical clustering and cluster-wise local ensembling, \ac{RMB-CLE} moves
beyond the binary inlier-outlier paradigm and provides a systematic
mechanism for discovering multiple task groups under heterogeneity.

From a methodological standpoint, the main contribution of \ac{RMB-CLE} lies in
its structural treatment of negative transfer. Rather than addressing
multi-task generalization through soft weighting or gradient manipulation
within a shared model, the proposed framework separates incompatible tasks
while preserving effective knowledge sharing among related ones. This is
achieved through three components: (i) cross-task error evaluation as a
reference for functional similarity, (ii) unsupervised discovery of latent task
clusters without metadata or predefined assumptions, and (iii) cluster-specific
ensemble learning that balances robustness and specialization. Moreover, from a
theoretical standpoint, \ac{RMB-CLE} is grounded in the fact that cross-task
generalization error provides a justified measure of task compatibility. In
regression, cross-task risk decomposes into a functional mismatch term and an
irreducible noise component, while in classification it upper-bounds the excess
risk induced by transferring a model across tasks.

Extensive empirical results on both synthetic and real-world benchmarks
demonstrate that \ac{RMB-CLE} consistently improves predictive performance
across classification and regression settings. On synthetic data, the framework
reliably recovers ground-truth task clusters and achieves performance
indistinguishable from oracle cluster-known baselines. On real-world datasets,
\ac{RMB-CLE} outperforms single-task learning, naive pooling, and existing
multi-task boosting methods, while maintaining stable performance under varying
numbers of tasks, class imbalance, and distributional shift. Task-wise
statistical tests further confirm that these gains are not driven by a small
subset of tasks but reflect systematic improvements.

Ablation studies, reported in the appendices, reinforce the central design
choices of the framework. In particular, clustering based on cross-task errors
consistently outperforms alternatives based on pseudo-residuals, highlighting
the importance of capturing long-run transferability rather than local
optimization dynamics. Moreover, the method is robust to the choice of
hierarchical linkage criterion, and silhouette-based model selection eliminates
the need for manual specification of the number of task clusters.

In summary, \ac{RMB-CLE} establishes a general and scalable principle for
robust multi-task learning with ensembles: task structure should be inferred
from how models transfer across tasks, rather than from how well they perform
in isolation. While the proposed framework is effective for moderate numbers of
tasks, its reliance on cross-task evaluation introduces a quadratic dependency
on the number of tasks, which motivates the development of more scalable
approximations for large-scale settings. In addition, the current formulation
assumes a common input dimension across tasks and does not address
heterogeneity in task dimensionality, which may arise in applications with
partially overlapping or task-specific feature spaces. Another promising
direction is to refine cross-task similarity estimation by accounting for task
difficulty, for example by normalizing transfer errors relative to the
intrinsic difficulty of the source and target tasks. Addressing these
limitations opens several promising directions for future work.

\section*{Declaration of competing interest}
The authors report no conflicts of interest or personal
relationships that may have influenced the research,
experimental work, or the conclusions of this study.

\section*{Acknowledgments}
\sloppy
The authors acknowledge financial support from project PID2022-139856NB-I00
(MCIN/AEI/10.13039/501100011033 and FEDER, UE), project IDEA-CM
(TEC-2024/COM-89, Comunidad de Madrid), and the ELLIS Unit Madrid, as well as
computational resources from the Centro de Computación Científica, Universidad
Autónoma de Madrid (CCC-UAM).

\section*{Data and Code availability}
The datasets used in this study were obtained from the
publicly available sources cited in the references.
Both the datasets and the source code developed for
the proposed model are publicly accessible at
\url{github.com/GAA-UAM/RMB-CLE}.
The generated synthetic datasets are also archived and publicly accessible via Mendeley Data at
\url{data.mendeley.com/datasets/3gms2fvs93/}.

\bibliographystyle{unsrtnat}
\bibliography{references}

\clearpage
\appendix

\section{Ablation study}~\label{appendix:ablation_study}
This ablation study analyzes the main design choices of \ac{RMB-CLE}. First, we
examine the impact of the task similarity definition by comparing cross-task
error-based similarity with alternative approach
(\ref{appendix:alternative_approach}). Second, we study the effect of the
clustering strategy, including different linkage criteria in hierarchical
clustering (\ref{appendix:alt_linkage}). Third, we investigate the use of
pseudo-residual-based similarity matrices as an alternative to cross-task error
evaluation (\ref{appendix:pseudo_residual_effects}).

\subsection{Alternative out-of-sample approach}~\label{appendix:alternative_approach}
An alternative to our cross-task error framework would be to introduce an
\emph{out-of-sample} validation procedure within each task. Formally, for every
task $i \in \mathcal{T}$ with dataset $\mathcal{D}_{i} = \{ (\mathbf{x}_{i,v},
      y_{i,v}) \}_{v=1}^{n_i}$, the samples could be partitioned into a training set
$\mathcal{D}_{i}^{\mathrm{train}}$ and an out-of-sample set
$\mathcal{D}_{i}^{\mathrm{out}}$. A task-specific predictor $F_{i}$ is then
trained on $\mathcal{D}_{i}^{\mathrm{train}}$ and evaluated on
$\mathcal{D}_{i}^{\mathrm{out}}$, producing residual vectors,
\[
      error_{i}^{\mathrm{out}} =
      \left\{\, y_{i,v} - F_{i}(\mathbf{x}_{i,v})
      \,\right\}_{v \in \mathcal{D}_{i}^{\mathrm{out}}}.
\]
One could then normalize these residuals and compute a similarity or distance
matrix across tasks based on $\{error_{i}^{\mathrm{out}}\}_{i=1}^{m}$, which in
turn could be used for clustering tasks.

Although such an approach appears natural, we deliberately chose \emph{not} to
adopt it for two main reasons:

\begin{enumerate}
      \item \textbf{Data efficiency:} In many tasks, especially
            those with small
            sample sizes, splitting into training and out-of-sample
            subsets reduces the
            effective sample size used for both model fitting and residual
            construction. This leads to instability in the estimated
            residuals
            $error_{i}^{\mathrm{out}}$.

      \item \textbf{Comparability across tasks:} Out-of-sample
            residuals rely on
            task-specific holdout splits; so, $error_i^{\mathrm{out}}$
            and $error_{j}^{\mathrm{out}}$
            are defined on different sample supports,
            with split-specific, task-dependent noise reflected
            in cross-task distances.
\end{enumerate}

\subsection{Effect of clustering linkage criterion}~\label{appendix:alt_linkage}
We repeated all synthetic experiments from Subsection~\ref{subsec:synthetic}
using Complete-linkage clustering instead of Average-linkage (UPGMA) to examine
whether the choice of linkage criterion affects the formation of task groups.
The results remained unchanged. In the regression setting, both linkage methods
using \ac{LGBM} produced identical performance, with \ac{RMSE} = $ 0.278$ $\pm$
$0.017$ and MAE = $0.218$ $\pm$ $0.013$. Likewise, in the classification
setting, accuracy and recall were exactly the same under both linkage
strategies.

These consistent outcomes indicate that the proposed framework is robust to the
choice of hierarchical clustering linkage. This stability arises because the
high-level task-group representations do not fluctuate across linkage criteria,
and the silhouette-based model selection
(Eqs.~\eqref{eq:silhouette}--\eqref{eq:bestk}) automatically selects the most
coherent clustering. Consequently, the method does not rely on manually
specifying the number of clusters and remains invariant to linkage variations.

\subsection{Effect of the pseudo-residual}~\label{appendix:pseudo_residual_effects}
The proposed \ac{RMB-CLE} framework is sensitive to the choice of the input
similarity matrix. Rather than computing cross-task errors, one can use
pseudo-residuals, which are defined as the \emph{negative gradient} of the loss
function ($\ell$) with respect to the model output ($F_{i,(t-1)}$) (See
Eq.~\eqref{eq:lgbm-recursion}). Formally, for each training instance of $v$ of
the task $i$, the pseudo-residual is given by,
\begin{equation}
      \rho_{iv, (t)}=
      -\left.
      \frac{\partial \ell\big(y_{i,v},\,F_{i, (t-1)}(\mathbf{x}_{i,v})\big)}
      {\partial F_{i, (t-1)}(\mathbf{x}_{i,v})}
      \right|_{F_{i, (t-1)}}.
      \label{eq:neg-gradient}
\end{equation}

In this subsection we report results obtained when pseudo-residuals are used to
build the similarity matrix (Eq.~\eqref{eq:similarity_matrix}) and we evaluate
performance on synthetic (\ref{subsec:spendix_synthetic}) and real-world
datasets (\ref{subsec:apendix_real_world}).

\subsubsection{Synthetic experiments}~\label{subsec:spendix_synthetic}
Following the setup described in Section~\ref{subsec:datasets}, we generated
another 100 repetitions of a synthetic dataset, each containing 25 tasks evenly
organized into 5 clusters, constructed using $\omega = 0.9$. Each task includes
300 training samples and 1{,}000 test samples with five input features,
evaluated under both classification and regression settings. The studied models
were trained and evaluated on these datasets, with \ac{RMB-CLE} constructing
the similarity matrix using pseudo-residuals for clustering.

Table~\ref{tab:synthetic_clf_performance_pseudo} presents the performance of
the studied models in the synthetic experiment. Comparing these results with
Tables~\ref{tab:synthetic_acc_rmse} and~\ref{tab:synthetic_recall_mae}, which
use cross-task errors for the similarity matrix, we observe that \ac{RMB-CLE}
is less effective here than when cross-task errors are used for both
classification and regression tasks.

\begin{table}[ht]
      \centering
      \small
      \caption{Synthetic multi-task results (mean ± Std Dev), analogous
            to Tables~\ref{tab:synthetic_acc_rmse} and~\ref{tab:synthetic_recall_mae},
            using pseudo-residuals instead of cross-task
            errors to construct the similarity matrix.}~\label{tab:synthetic_clf_performance_pseudo}
      \resizebox{1.0\textwidth}{!}{%
            \begin{tabular}{lcccccc}
                  \toprule
                  \textbf{Model}     & \textbf{Accuracy}      & \textbf{Recall}        & \textbf{RMSE}          & \textbf{MAE}           \\
                  \midrule
                  RMB-CLE-via-LGBM   & \textbf{0.873 ± 0.019} & \textbf{0.873 ± 0.019} & 0.338 ± 0.020          & 0.266 ± 0.015          \\
                  Cluster-Known-LGBM & \textbf{0.873 ± 0.019} & \textbf{0.873 ± 0.019} & 0.338 ± 0.020          & 0.266 ± 0.015          \\
                  Cluster-Known-MTGB & 0.818 ± 0.015          & 0.818 ± 0.015          & 0.336 ± 0.020          & 0.264 ± 0.016          \\
                  RMB-CLE-via-MTGB   & 0.818 ± 0.015          & 0.818 ± 0.015          & 0.336 ± 0.020          & 0.264 ± 0.016          \\
                  ST-LGBM            & 0.804 ± 0.016          & 0.804 ± 0.016          & \textbf{0.328 ± 0.013} & \textbf{0.258 ± 0.010} \\
                  ST-GB              & 0.803 ± 0.016          & 0.803 ± 0.016          & 0.357 ± 0.015          & 0.280 ± 0.012          \\
                  R-MTGB             & 0.733 ± 0.052          & 0.733 ± 0.052          & 0.540 ± 0.103          & 0.431 ± 0.086          \\
                  MTGB               & 0.728 ± 0.084          & 0.728 ± 0.084          & 0.792 ± 0.203          & 0.644 ± 0.172          \\
                  DP-GB              & 0.628 ± 0.031          & 0.628 ± 0.031          & 0.842 ± 0.157          & 0.686 ± 0.135          \\
                  TaF-GB             & 0.628 ± 0.031          & 0.628 ± 0.031          & 0.547 ± 0.044          & 0.435 ± 0.035          \\
                  DP-LGBM            & 0.623 ± 0.031          & 0.623 ± 0.031          & 0.840 ± 0.157          & 0.685 ± 0.135          \\
                  TaF-LGBM           & 0.622 ± 0.031          & 0.622 ± 0.031          & 0.545 ± 0.044          & 0.434 ± 0.035          \\
                  \bottomrule
            \end{tabular}
      }
\end{table}

\begin{figure}[ht]
      \centering
      \includegraphics[width=1.0\linewidth]{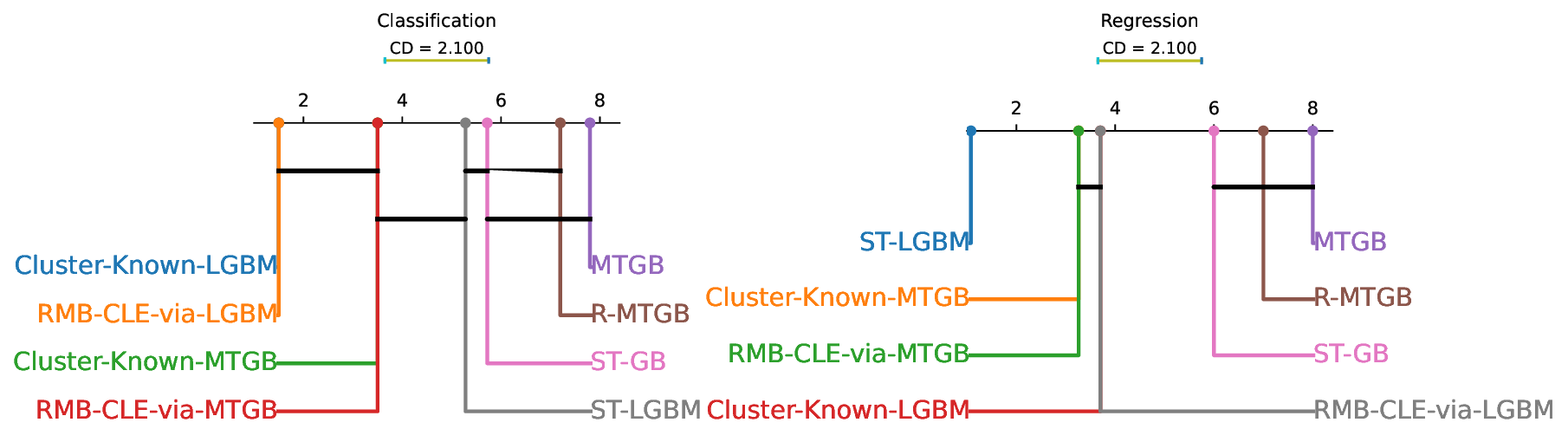}
      \caption{Synthetic task-wise Dem\v{s}ar plots
            (p = 0.05), analogous to Figure~\ref{fig:Figure_2}, where task
            clustering is performed using pseudo-residuals instead of
            cross-task errors. Rankings are computed across
            classification and regression tasks.}
      \label{fig:Figure_7}
\end{figure}

The per-task performance of the task-wise models is presented in
Figure~\ref{fig:Figure_7}, shown as a Dems\v{a}r ranking plot with statistical
differences measured using the Nemenyi test. The proposed framework, yet
clustered using pseudo-residuals consistently achieved the best or second-best
rankings in classification problem (left subplot) and the third and fourth
positions in regression problem (right subplot). In contrast, conventional
boosting-based multi-task models (\ac{R-MTGB} and \ac{MTGB}) obtained the
lowest rankings across all problems (subplots).

\subsubsection{Real-world experiments}~\label{subsec:apendix_real_world}
Tables~\ref{tab:realworld_clf_performance_accuracy_pseudo}
and~\ref{tab:realworld_reg_performance_rmse_pseudo} report real-world results
for the proposed \ac{RMB-CLE} framework when task clustering is performed using
pseudo-residuals (Eq.~\ref{eq:neg-gradient}) rather than cross-task errors (Eq.~\ref{eq:cross-error}).
Following the same evaluation protocol as in the real-world experiments
(Subsection~\ref{subsec:realworld}), each experiment is repeated 100 times; in
each run, performance is first averaged across tasks, and the final results are
reported as the mean and \ac{Std Dev} over the 100 runs

Compared to clustering based on cross-task errors, the pseudo-residual variant
achieves less stable improvements. While \ac{RMB-CLE} with pseudo-residual
clustering can still match or occasionally outperform certain baselines (e.g.,
on \textit{Adult} and \textit{SARCOS}), its gains are less consistent and, in
several cases, inferior to those obtained with the cross-task error-based
similarity.

\begin{table}[ht]
      \centering
      \small
      \caption{Real-world multi-task classification accuracy
            (mean ± Std Dev),
            analogous to Table~\ref{tab:realworld_clf_performance},
            using pseudo-residual-based clustering.}
      \label{tab:realworld_clf_performance_accuracy_pseudo}
      \resizebox{1.0\textwidth}{!}{%
            \begin{tabular}{lccccc}
                  \toprule
                  \textbf{Model}   & \textbf{Adult-Gender}      & \textbf{Adult-Race}        & \textbf{Avila}                      & \textbf{Bank}              & \textbf{Landmine}          \\
                  \midrule
                  DP-GB            & 0.837 $\pm$ 0.005          & 0.837 $\pm$ 0.005          & 0.494 $\pm$ 0.010                   & 0.889 $\pm$ 0.003          & 0.939 $\pm$ 0.003          \\
                  DP-LGBM          & 0.854 $\pm$ 0.004          & 0.854 $\pm$ 0.004          & 0.536 $\pm$ 0.021                   & 0.898 $\pm$ 0.003          & 0.939 $\pm$ 0.003          \\
                  RMB-CLE-via-LGBM & \textbf{0.872 $\pm$ 0.003} & \textbf{0.873 $\pm$ 0.003} & \textbf{0.872} $\pm$ \textbf{0.150} & \textbf{0.902 $\pm$ 0.003} & 0.941 $\pm$ 0.004          \\
                  RMB-CLE-via-MTGB & 0.848 $\pm$ 0.004          & 0.838 $\pm$ 0.005          & 0.604 $\pm$ 0.055                   & 0.894 $\pm$ 0.003          & 0.943 $\pm$ 0.004          \\
                  MTGB             & 0.848 $\pm$ 0.004          & 0.845 $\pm$ 0.004          & 0.614 $\pm$ 0.050                   & 0.893 $\pm$ 0.003          & 0.943 $\pm$ 0.004          \\
                  R-MTGB           & 0.849 $\pm$ 0.004          & 0.849 $\pm$ 0.004          & 0.619 $\pm$ 0.047                   & 0.895 $\pm$ 0.003          & \textbf{0.943 $\pm$ 0.004} \\
                  ST-GB            & 0.841 $\pm$ 0.004          & 0.838 $\pm$ 0.004          & 0.610 $\pm$ 0.060                   & 0.892 $\pm$ 0.003          & 0.942 $\pm$ 0.003          \\
                  ST-LGBM          & 0.855 $\pm$ 0.003          & 0.852 $\pm$ 0.004          & 0.690 $\pm$ 0.092                   & 0.897 $\pm$ 0.003          & 0.940 $\pm$ 0.003          \\
                  TaF-GB           & 0.837 $\pm$ 0.005          & 0.837 $\pm$ 0.005          & 0.497 $\pm$ 0.009                   & 0.889 $\pm$ 0.003          & 0.939 $\pm$ 0.003          \\
                  TaF-LGBM         & 0.854 $\pm$ 0.004          & 0.854 $\pm$ 0.004          & 0.552 $\pm$ 0.114                   & 0.898 $\pm$ 0.003          & 0.939 $\pm$ 0.003          \\
                  \bottomrule
            \end{tabular}
      }
\end{table}

\begin{table}[ht]
      \centering
      \small
      \caption{Real-world multi-task regression RMSE
            (mean ± Std Dev),
            analogous to Table~\ref{tab:realworld_reg_performance},
            using pseudo-residual-based clustering.}
      \label{tab:realworld_reg_performance_rmse_pseudo}
      \resizebox{1.0\textwidth}{!}{%
            \begin{tabular}{lccccc}
                  \toprule
                  \textbf{Model}   & \textbf{Abalone}           & \textbf{Computer}          & \textbf{Parkinson}         & \textbf{SARCOS}                     & \textbf{School}                      \\
                  \midrule
                  DP-GB            & 2.397 $\pm$ 0.093          & 2.466 $\pm$ 0.048          & 8.859 $\pm$ 0.137          & 18.397 $\pm$ 0.067                  & 10.423 $\pm$ 0.118                   \\
                  DP-LGBM          & 2.403 $\pm$ 0.093          & 2.466 $\pm$ 0.048          & 8.859 $\pm$ 0.136          & 18.395 $\pm$ 0.067                  & 10.424 $\pm$ 0.118                   \\
                  RMB-CLE-via-LGBM & \textbf{2.169 $\pm$ 0.085} & 2.933 $\pm$ 0.221          & 0.495 $\pm$ 0.039          & \textbf{2.203} $\pm$ \textbf{0.016} & 10.725 $\pm$ 0.128                   \\
                  RMB-CLE-via-MTGB & 2.465 $\pm$ 0.081          & 2.754 $\pm$ 0.333          & \textbf{0.209 $\pm$ 0.029} & 7.561 $\pm$ 0.051                   & 10.292 $\pm$ 0.139                   \\
                  MTGB             & 2.289 $\pm$ 0.087          & 2.486 $\pm$ 0.047          & 0.336 $\pm$ 0.024          & 4.808 $\pm$ 0.034                   & 10.154 $\pm$ 0.122                   \\
                  R-MTGB           & 2.266 $\pm$ 0.086          & \textbf{2.463 $\pm$ 0.076} & 0.289 $\pm$ 0.037          & 4.698 $\pm$ 0.066                   & \textbf{10.133} $\pm$ \textbf{0.119} \\
                  ST-GB            & 2.346 $\pm$ 0.089          & 2.760 $\pm$ 0.352          & 0.268 $\pm$ 0.027          & 4.919 $\pm$ 0.034                   & 10.295 $\pm$ 0.137                   \\
                  ST-LGBM          & 2.345 $\pm$ 0.089          & 2.948 $\pm$ 0.214          & 0.557 $\pm$ 0.035          & 4.926 $\pm$ 0.034                   & 10.669 $\pm$ 0.131                   \\
                  TaF-GB           & 2.383 $\pm$ 0.093          & 2.467 $\pm$ 0.067          & 6.559 $\pm$ 0.087          & 11.266 $\pm$ 0.060                  & 10.415 $\pm$ 0.117                   \\
                  TaF-LGBM         & 2.388 $\pm$ 0.093          & 2.469 $\pm$ 0.048          & 6.558 $\pm$ 0.087          & 11.266 $\pm$ 0.060                  & 10.416 $\pm$ 0.118                   \\
                  \bottomrule
            \end{tabular}
      }
\end{table}

\begin{figure}[ht]
      \centering
      \includegraphics[width=1.0\linewidth]{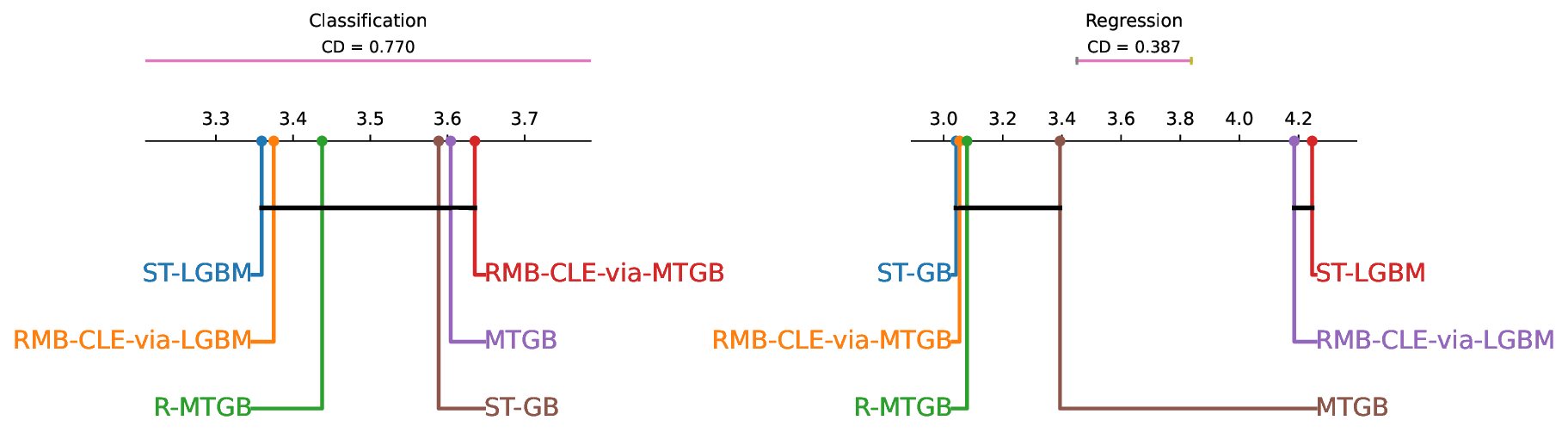}
      \caption{Real-world task-wise Dem\v{s}ar plots (p = 0.05),
            analogous to Figure~\ref{fig:Figure_6}, with
            task clustering based on pseudo-residuals rather
            than cross-task errors.}
      \label{fig:Figure_8}
\end{figure}

The related task-wise rankings (Figure~\ref{fig:Figure_8}) confirm the results
reported in Tables~\ref{tab:realworld_clf_performance_accuracy_pseudo}
and~\ref{tab:realworld_reg_performance_rmse_pseudo} showing that using
pseudo-residuals instead of cross-task errors to construct the similarity
matrix does not guarantee that the model will be optimal or achieve better
performance.

These observations confirm that the quality of the clustering signal is
critical: cross-task errors generally provide a more reliable basis for task
grouping than pseudo-residuals. Although the pseudo-residual variant of
\ac{RMB-CLE} remains competitive in some settings, it cannot be considered the
best-performing approach overall, and its usefulness should be carefully
weighed against the residual-based alternative.

\subsubsection{Discussion and limitations}
While clustering based on pseudo-residuals could returns partitions that appear
similar to those obtained from cross-task errors, the resulting performance is
generally inferior. The reason is that pseudo-residuals capture only local
optimization dynamics (gradients of the loss at a given training iteration)
rather than long-run transferability between tasks. Consequently, the
similarity profiles they generate are less stable, more sensitive to training
noise, and do not consistently reflect functional relatedness. Even if the
final task assignments were to coincide with those from cross-task errors, the
underlying similarity geometry could still differ. For illustration,
pseudo-residual similarities might flatten the relative closeness of tasks
(e.g., treating $m_1 \approx m_2 \approx m_3$ instead of the true $m_1 \approx
      m_2 < m_3$), so that clusters appear correct at the partition level but do not
reflect genuine functional relationships. Cross-task errors, by contrast,
directly evaluate out-of-domain generalization by testing models trained on one
task against another.

One might also consider computing pseudo-residuals in a cross-task fashion
(i.e., evaluating residuals from task $j$ on task $i$). However, this would
largely duplicate the information already contained in cross-task errors, while
adding instability due to the iteration-dependent nature of residuals and the
arbitrary need to select training iterations. Moreover, the clustering signal
obtained in this way would still be less reliable, as pseudo-residuals
emphasize gradient noise rather than functional similarity. For these reasons,
and given the consistent empirical advantage of error-based similarities across
our ablations, we restrict the main \ac{RMB-CLE} framework to clustering with
cross-task errors.

\section{Additional experiments and results}~\label{appendix:additional_experiments}
This appendix presents additional experimental analyses
that extend the results discussed in
Section~\ref{sec:experiments}. First, in~\ref{apendix:time_cost}, we examine the
computational costs of the models and approaches
under study. Next, we complement this analysis
with empirical measurements of elapsed
runtime across different scenarios
(\ref{apendix:time_tables}).
Finally, in~\ref{apendix:F1_score}, we report
an additional accuracy metric, the F1-score, to provide
insight into the experimental results under class imbalance.

\subsection{Computational Cost}~\label{apendix:time_cost}
This section analyzes the computational cost of the proposed \ac{RMB-CLE}
framework and compares its asymptotic complexity with \ac{GB}, \ac{MTGB}, and
\ac{R-MTGB}. We express
computational cost in terms of boosting blocks ($\mathcal{S}_{(.)}$), the number of tasks
($m$), and the average number of training instances per task ($\bar{n}$). For
each methodology/approach, we report both training and latency complexities
using Big-$\mathcal{O}$ notation.

Different boosting blocks $\mathcal{S}_{(.)}$ are distinguished by their
functional role within the learning architecture rather than by their internal
optimization mechanism. While all blocks consist of sequential boosting
iterations trained in the same manner, their indices indicate the type of
information they model and the scope over which they operate. For instance,
blocks may correspond to shared learning across tasks ($\mathcal{S}_{(1)}$),
task-specific refinement ($\mathcal{S}_{(2)}$), outlier-aware weighting
($\mathcal{S}_{(3)}$), or cluster-level ensembling ($\mathcal{S}_{(4)}$).

\subsubsection*{Training complexity}

\paragraph{Single-Task approaches}
Standard \ac{ST} approach consists of a single boosting block of
$\mathcal{K}_{\mathcal{S}_{(2)}}$ base learners. Its training complexity is
\[
      \mathcal{O}(\mathcal{K}_{\mathcal{S}_{(2)}} \, \bar{n}).
\]

\paragraph{MTGB}
\ac{MTGB} is composed of two sequential boosting blocks: a shared block
of $\mathcal{K}_{\mathcal{S}_{(1)}}$ base
learners
and a
task-specific block of $\mathcal{K}_{\mathcal{S}_{(2)}}$ base
learners. Hence, the training complexity scales as
\[
      \mathcal{O}\big((\mathcal{K}_{\mathcal{S}_{(1)}} + \mathcal{K}_{\mathcal{S}_{(2)}})\, m \, \bar{n}\big).
\]

\paragraph{R-MTGB}
This model extends \ac{MTGB} by introducing an additional outlier-aware block
$\mathcal{S}_{(3)}$ between the shared and task-specific blocks, resulting in
three sequential blocks of $\mathcal{K}_{\mathcal{S}_{(1)}}$,
$\mathcal{K}_{\mathcal{S}_{(2)}}$, and $\mathcal{K}_{\mathcal{S}_{(3)}}$ base
learners, with two branches in the third block. Therefore, its training
complexity is
\[
      \mathcal{O}\big((\mathcal{K}_{\mathcal{S}_{(1)}} +
      \mathcal{K}_{\mathcal{S}_{(2)}} +
      2\mathcal{K}_{\mathcal{S}_{(3)}})\, m \, \bar{n}\big),
\]
which has the same asymptotic order as \ac{MTGB} but with a larger constant
factor~\cite{Emami2025}.

\paragraph{RMB-CLE}
The proposed framework follows a different strategy by evaluating task
similarity and forming task clusters prior to cluster-wise ensemble training.

\begin{itemize}
      \item \textbf{Initial task-wise boosting:} each task is first modeled
            independently using task-specific boosting block of
            $\mathcal{K}_{\mathcal{S}_{(2)}}$ predictors
            resulting in
            \[
                  \mathcal{O}(\mathcal{K}_{\mathcal{S}_{(2)}} \, m \, \bar{n}).
            \]

      \item \textbf{Cross-task evaluation:} each task model is evaluated on all other
            tasks to construct the cross-task error matrix, incurring
            \[
                  \mathcal{O}(m^2 \, \bar{n}).
            \]

      \item \textbf{Similarity computation and clustering:} building the similarity
            matrix and performing hierarchical clustering costs
            \[
                  \mathcal{O}(m^2 \log m).
            \]

      \item \textbf{Cluster-wise ensemble training:} For each identified cluster,
            a local boosting ensemble of $\mathcal{K}_{\mathcal{S}_{(4)}}$  base
            learners is trained
            using the pooled data within that cluster.
            Aggregating over all clusters,
            the computational complexity of this step is
            \[
                  \mathcal{O}\!\left(\mathcal{K}_{\mathcal{S}_{(4)}} \, m \, \bar{n}\right).
            \]
\end{itemize}

The total training complexity of RMB-CLE is
\[
      \mathcal{O}\big((\mathcal{K}_{\mathcal{S}_{(2)}} +
      \mathcal{K}_{\mathcal{S}_{(4)}})\, m \, \bar{n}\big)
      \;+\;
      \mathcal{O}(m^2 \, \bar{n})
      \;+\;
      \mathcal{O}(m^2 \log m).
\]

\subsubsection*{Latency complexity}

At prediction time, prediction for a single test instance requires evaluating
the sequence of boosting blocks associated with the corresponding task. Since
the number of boosting iterations within each block is fixed after training,
latency cost is independent of the number of tasks $m$ and the average number
of training samples $\bar{n}$. However, different methods have different
constant-time costs depending on the number of boosting blocks and base
learners that must be evaluated.

\paragraph{Single-task approaches}
Prediction requires evaluating a single boosting block consisting of
$\mathcal{K}_{\mathcal{S}_{(2)}}$ base learners, resulting in a per-instance
latency of
\[
      \mathcal{O}\!\left(\mathcal{K}_{\mathcal{S}_{(2)}}\right).
\]

\paragraph{MTGB}
Prediction involves evaluating both the shared and task-specific boosting
blocks, with $\mathcal{K}_{\mathcal{S}_{(1)}}$ and
$\mathcal{K}_{\mathcal{S}_{(2)}}$ base learners, respectively. The resulting
per-instance latency is
\[
      \mathcal{O}\!\left(\mathcal{K}_{\mathcal{S}_{(1)}} + \mathcal{K}_{\mathcal{S}_{(2)}}\right).
\]

\paragraph{R-MTGB}
This model evaluates a shared block, an outlier-aware block with two branches,
and a task-specific block. Consequently, prediction requires
\[
      \mathcal{O}\!\left(\mathcal{K}_{\mathcal{S}_{(1)}} + \mathcal{K}_{\mathcal{S}_{(2)}} + 2\mathcal{K}_{\mathcal{S}_{(3)}}\right),
\]
operations per test instance, showing a larger constant factor than \ac{MTGB}.

\paragraph{RMB-CLE}
Given a task identifier, prediction consists of a constant-time cluster lookup
followed by evaluation of the corresponding cluster-level ensemble with
$\mathcal{K}_{\mathcal{S}_{(4)}}$ base learners. Therefore, the per-instance
latency of is
\[
      \mathcal{O}\!\left(\mathcal{K}_{\mathcal{S}_{(4)}}\right).
\]

A summary of the training and latency complexities of the studied methods and
approaches is reported in Table~\ref{tab:complexity_summary}.

\begin{table}[t]
      \centering
      \caption{Summary of training and latency complexity for the studied methods.}
      \label{tab:complexity_summary}
      \resizebox{1.0\textwidth}{!}{%
            \begin{tabular}{lll}
                  \toprule
                  \textbf{Method/Approach}                                & \textbf{Training Complexity}                                                                                       & \textbf{Inference Complexity} \\
                  \midrule
                  Single-Task
                                                                          & $\mathcal{O}\!\left(\mathcal{K}_{\mathcal{S}_{(2)}} \, \bar{n}\right)$
                                                                          & $\mathcal{O}\!\left(\mathcal{K}_{\mathcal{S}_{(2)}}\right)$                                                                                        \\[6pt]

                  MTGB                                                    & $\mathcal{O}\!\left((\mathcal{K}_{\mathcal{S}_{(1)}} +
                  \mathcal{K}_{\mathcal{S}_{(2)}})\, m \, \bar{n}\right)$ &
                  $\mathcal{O}\!\left(\mathcal{K}_{\mathcal{S}_{(1)}} +
                  \mathcal{K}_{\mathcal{S}_{(2)}}\right)$                                                                                                                                                                      \\[6pt]

                  R-MTGB                                                  & $\mathcal{O}\!\left((\mathcal{K}_{\mathcal{S}_{(1)}} +
                        \mathcal{K}_{\mathcal{S}_{(2)}} + 2\mathcal{K}_{\mathcal{S}_{(3)}})\, m \,
                  \bar{n}\right)$                                         & $\mathcal{O}\!\left(\mathcal{K}_{\mathcal{S}_{(1)}} +
                  \mathcal{K}_{\mathcal{S}_{(2)}} + 2\mathcal{K}_{\mathcal{S}_{(3)}}\right)$                                                                                                                                   \\[6pt]

                  RMB-CLE                                                 & $\begin{aligned}
                                                                                          & \mathcal{O}\!\left((\mathcal{K}_{\mathcal{S}_{(2)}} + \mathcal{K}_{\mathcal{S}_{(4)}})\, m \, \bar{n}\right) \\
                                                                                          & \quad + \mathcal{O}(m^2 \, \bar{n}) + \mathcal{O}(m^2 \log m)
                                                                                   \end{aligned}$
                                                                          & $\mathcal{O}\!\left(\mathcal{K}_{\mathcal{S}_{(4)}}\right)$                                                                                        \\

                  \bottomrule
            \end{tabular}
      }
\end{table}

\subsection{Time evaluation}~\label{apendix:time_tables}
For each batch of synthetic experiments, we measured elapsed \emph{wall-clock}
time (in seconds) using \texttt{time.perf\_counter} of \texttt{Python}, a
high-resolution timer designed for precise performance measurements.
For each model configuration, we decomposed the runtime into four components:
(i) \emph{grid-search time}, defined as the time required to run the
\texttt{GridSearch} procedure with cross-validation;
(ii) \emph{best-fit time}, corresponding to the time needed to train a single
model on the full training set using the hyperparameter configuration selected
by the grid search;
(iii) \emph{total training time}, computed as the sum of the
\emph{grid-search time} and the \emph{best-fit time};
and (iv) \emph{prediction time}, measured as the time required to generate
predictions on the test split using the fitted best model.
We repeated this procedure over 100 experimental runs (corresponding to 100 independently generated synthetic datasets) and report the mean and
\ac{Std Dev} of the per-batch times in Tables~\ref{tab:time_clf}
and~\ref{tab:time_reg} for the classification and regression synthetic
datasets, respectively. The best (fastest) result in each column is highlighted
in \textbf{bold}.

Across both settings, the relative ordering of methods is highly consistent,
showing that computational behavior is primarily driven by model architecture
rather than problem type. Pooling and single-task baselines, particularly
\ac{DP}-\ac{LGBM}, achieves the lowest runtimes across all components, as they
train a single global model (\ac{DP}) or independent task-wise models (\ac{ST})
without multi-stage or cluster-specific structures. \ac{TaF} models experience
slightly higher costs due to the augmented input space but remain
computationally efficient since they still rely on a single shared ensemble.

In contrast, \ac{MT} boosting methods such as \ac{MTGB} and \ac{R-MTGB} exhibit
higher grid-search and training times, reflecting their sequential multi-block
architectures. The \ac{MTGB} trains separate shared and task-specific boosting
stages, while \ac{R-MTGB} further introduces an outlier-aware block, increasing
both the number of boosting stages and the dimensionality of the hyperparameter
search. Cluster-based approaches, including the oracle Cluster-Known baselines
and the proposed \ac{RMB-CLE} variants, show nearly identical timing profiles
when instantiated with the same local ensemble ($f\in\mathcal{F}$ in
Eq.~\eqref{eq:local-ensemble}). This is because, once clusters are fixed, both
methods reduce to training cluster-specific ensembles on pooled data, and the
clustering step itself introduces negligible overhead. Across all methods,
configurations based on \ac{LGBM} consistently achieve lower training times due
to the histogram-based feature binning and shallow tree construction in
\ac{LGBM}, which reduce split-search complexity, memory access costs, and
per-iteration computation compared to standard \ac{GB} implementations.
Furthermore, \ac{RMB-CLE} avoids the sequential per-task boosting structure of
\ac{MTGB} and \ac{R-MTGB} (when \ac{LGBM} is used as the local ensemble),
resulting in lower training times while maintaining comparable prediction
latency, which remains low due to constant-time cluster lookup followed by a
single ensemble evaluation.

\begin{table}[ht]
      \centering
      \caption{Classification wall-clock time in
            seconds (mean ± Std Dev). Best results in bold.}~\label{tab:time_clf}
      \resizebox{1.0\textwidth}{!}{%
            \begin{tabular}{lcccc}
                  \toprule
                  \textbf{Model}               &
                  \textbf{Grid search time}    &
                  \textbf{Best fit time}       &
                  \textbf{Total training time} &
                  \textbf{Prediction time}                                                                                                                         \\
                  \midrule
                  Cluster-Known-LGBM           & 7.028 $\pm$ 0.595          & 3.374 $\pm$ 0.373          & 10.402 $\pm$ 0.702         & 0.193 $\pm$ 0.021          \\
                  Cluster-Known-MTGB           & 21.607 $\pm$ 1.720         & 4.927 $\pm$ 0.668          & 26.534 $\pm$ 1.845         & 0.458 $\pm$ 0.095          \\
                  DP-GB                        & 1.522 $\pm$ 0.140          & 0.757 $\pm$ 0.101          & 2.279 $\pm$ 0.173          & 0.160 $\pm$ 0.022          \\
                  DP-LGBM                      & \textbf{0.139 $\pm$ 0.015} & \textbf{0.045 $\pm$ 0.009} & \textbf{0.184 $\pm$ 0.018} & \textbf{0.057 $\pm$ 0.012} \\
                  RMB-CLE-via-LGBM             & 7.013 $\pm$ 0.597          & 3.374 $\pm$ 0.366          & 10.387 $\pm$ 0.700         & 0.193 $\pm$ 0.021          \\
                  RMB-CLE-via-MTGB             & 21.679 $\pm$ 1.815         & 4.924 $\pm$ 0.671          & 26.603 $\pm$ 1.935         & 0.458 $\pm$ 0.096          \\
                  MTGB                         & 8.589 $\pm$ 0.649          & 2.591 $\pm$ 0.255          & 11.180 $\pm$ 0.698         & 0.862 $\pm$ 0.050          \\
                  R-MTGB                       & 32.456 $\pm$ 2.194         & 3.542 $\pm$ 0.390          & 35.998 $\pm$ 2.229         & 0.892 $\pm$ 0.052          \\
                  ST-GB                        & 4.047 $\pm$ 0.366          & 1.434 $\pm$ 0.194          & 5.481 $\pm$ 0.414          & 0.200 $\pm$ 0.018          \\
                  ST-LGBM                      & 1.737 $\pm$ 0.088          & 0.504 $\pm$ 0.050          & 2.241 $\pm$ 0.101          & 0.070 $\pm$ 0.006          \\
                  TaF-GB                       & 1.824 $\pm$ 0.186          & 0.850 $\pm$ 0.134          & 2.674 $\pm$ 0.229          & 0.195 $\pm$ 0.043          \\
                  TaF-LGBM                     & 0.361 $\pm$ 0.031          & 0.103 $\pm$ 0.023          & 0.463 $\pm$ 0.039          & 0.074 $\pm$ 0.026          \\
                  \bottomrule
            \end{tabular}
      }
\end{table}

\begin{table}[ht]
      \centering
      \caption{Regression wall-clock time in
            seconds (mean ± Std Dev). Best results in bold.}~\label{tab:time_reg}
      \resizebox{1.0\textwidth}{!}{%
            \begin{tabular}{lcccc}
                  \toprule
                  \textbf{Model}     & \textbf{Grid search time}  & \textbf{Best fit time}     & \textbf{Total training time} & \textbf{Prediction time}   \\
                  \midrule
                  Cluster-Known-LGBM & 5.056 $\pm$ 0.342          & 2.424 $\pm$ 0.164          & 7.480 $\pm$ 0.379            & 0.210 $\pm$ 0.010          \\
                  Cluster-Known-MTGB & 14.735 $\pm$ 0.996         & 3.484 $\pm$ 0.235          & 18.219 $\pm$ 1.023           & 0.397 $\pm$ 0.035          \\
                  DP-GB              & 1.197 $\pm$ 0.083          & 0.598 $\pm$ 0.038          & 1.795 $\pm$ 0.091            & 0.086 $\pm$ 0.005          \\
                  DP-LGBM            & \textbf{0.119 $\pm$ 0.013} & \textbf{0.038 $\pm$ 0.002} & \textbf{0.157 $\pm$ 0.013}   & \textbf{0.062 $\pm$ 0.002} \\
                  RMB-CLE-via-LGBM   & 5.068 $\pm$ 0.338          & 2.427 $\pm$ 0.159          & 7.495 $\pm$ 0.374            & 0.210 $\pm$ 0.010          \\
                  RMB-CLE-via-MTGB   & 14.697 $\pm$ 1.006         & 3.484 $\pm$ 0.245          & 18.181 $\pm$ 1.035           & 0.396 $\pm$ 0.035          \\
                  MTGB               & 6.691 $\pm$ 0.424          & 2.065 $\pm$ 0.131          & 8.756 $\pm$ 0.444            & 0.609 $\pm$ 0.033          \\
                  R-MTGB             & 24.144 $\pm$ 1.637         & 2.754 $\pm$ 0.201          & 26.899 $\pm$ 1.649           & 0.635 $\pm$ 0.035          \\
                  ST-GB              & 3.382 $\pm$ 0.167          & 1.267 $\pm$ 0.063          & 4.649 $\pm$ 0.178            & 0.217 $\pm$ 0.013          \\
                  ST-LGBM            & 1.725 $\pm$ 0.094          & 0.558 $\pm$ 0.038          & 2.282 $\pm$ 0.102            & 0.081 $\pm$ 0.004          \\
                  TaF-GB             & 1.490 $\pm$ 0.118          & 0.694 $\pm$ 0.062          & 2.184 $\pm$ 0.133            & 0.118 $\pm$ 0.024          \\
                  TaF-LGBM           & 0.336 $\pm$ 0.017          & 0.098 $\pm$ 0.005          & 0.434 $\pm$ 0.018            & 0.086 $\pm$ 0.032          \\
                  \bottomrule
            \end{tabular}
      }
\end{table}

\begin{figure}[H]
      \centering
      \includegraphics[width=1.0\linewidth]{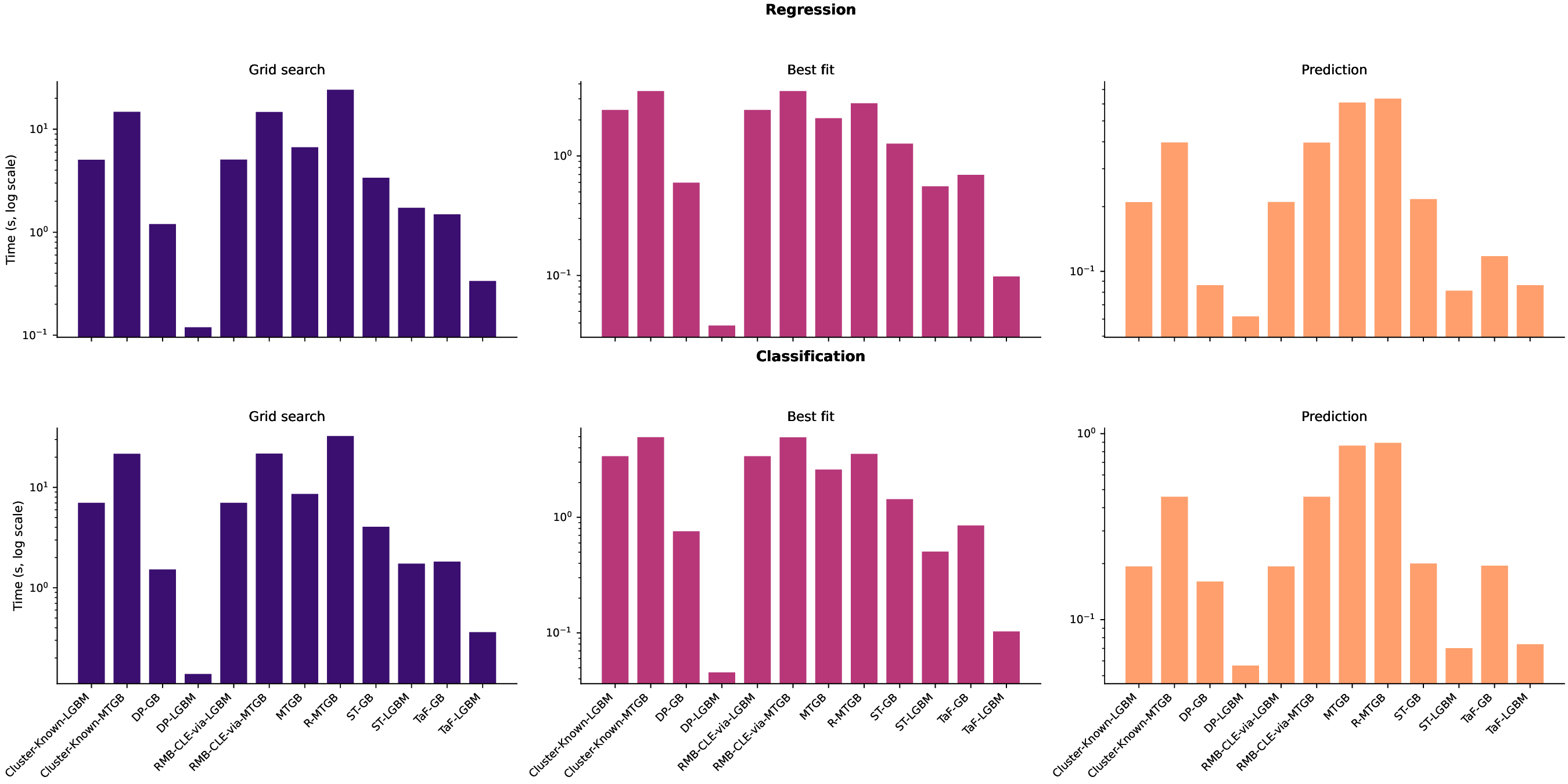}
      \caption{Runtime decomposition (log scale) for regression (top) and
            classification (bottom), showing mean grid-search, best-fit training, and prediction
            times over 100 runs for each model.}~\label{fig:Figure_5}
\end{figure}

While Tables~\ref{tab:time_clf} and~\ref{tab:time_reg} provide numerical
summaries, the substantial differences in magnitude between grid-search,
training, and prediction times make it difficult to assess their relative
contributions at a glance. To address this, we separate these components into
dedicated panels in Figure~\ref{fig:Figure_5}, which presents log-scale
visualizations of the mean runtime for regression (top) and classification
(bottom). Figure~\ref{fig:Figure_5} shows that grid-search and best-fit
training dominate the total runtime across all methods, with particularly high
costs for \ac{R-MTGB}. This behavior reflects its sequential block structure
and enlarged hyperparameter space. In contrast, \ac{DP}, \ac{ST}, and \ac{TaF}
baselines have lower mean elapsed time due to their simpler training
procedures. Prediction time is reported separately, as it is consistently one
to two orders of magnitude smaller than training-related costs across all
models and is largely insensitive to the underlying architecture. The
qualitative runtime patterns are nearly identical for classification and
regression.

\subsection{F1-score}~\label{apendix:F1_score}
In addition to accuracy and recall reported in the experiments described in
Subsection~\ref{subsec:realworld}, we also report the F1-score computed using
the macro averaging scheme. This provides a balanced evaluation of
classification performance under class imbalance and complements the results
presented in Tables~\ref{tab:realworld_clf_performance} and
\ref{tab:realworld_clf_performance_recall}.

Table~\ref{tab:f1} reports the mean and \ac{Std Dev} of the test F1-score over
100 runs for all real-world multi-task classification datasets. Results are
first averaged across tasks and then across repetitions, and the
best-performing method for each dataset is highlighted in \textbf{bold}.
Consistent with the accuracy and recall results presented in
Tables~\ref{tab:realworld_clf_performance}
and~\ref{tab:realworld_clf_performance_recall}, RMB-CLE-via-LGBM achieves the
highest F1-score across all datasets, indicating that its performance gains are
not driven by a single metric but reflect a robust improvement in the
precision-recall balance. In contrast, pooling-based and heuristic \ac{MT}
baselines exhibit substantially lower F1-scores, particularly on highly
imbalanced datasets such as \textit{Avila} and \textit{Landmine}.

\begin{table}[ht]
      \centering
      \small
      \caption{Real-world classification F1-score (mean ± Std Dev).
            Best results in bold.}
      \label{tab:f1}
      \resizebox{1.0\textwidth}{!}{%
            \begin{tabular}{lccccc}
                  \toprule
                  \textbf{Model}
                   & \textbf{Adult-Gender}
                   & \textbf{Adult-Race}
                   & \textbf{Avila}
                   & \textbf{Bank}
                   & \textbf{Landmine}          \\
                  \midrule
                  DP-GB
                   & 0.717 $\pm$ 0.011
                   & 0.717 $\pm$ 0.011
                   & 0.144 $\pm$ 0.006
                   & 0.532 $\pm$ 0.007
                   & 0.484 $\pm$ 0.001          \\
                  DP-LGBM
                   & 0.769 $\pm$ 0.005
                   & 0.769 $\pm$ 0.005
                   & 0.270 $\pm$ 0.028
                   & 0.646 $\pm$ 0.008
                   & 0.490 $\pm$ 0.007          \\
                  RMB-CLE-via-LGBM
                   & \textbf{0.814 $\pm$ 0.004}
                   & \textbf{0.815 $\pm$ 0.004}
                   & \textbf{0.772 $\pm$ 0.173}
                   & \textbf{0.738 $\pm$ 0.009}
                   & \textbf{0.635 $\pm$ 0.019} \\
                  RMB-CLE-via-MTGB
                   & 0.752 $\pm$ 0.006
                   & 0.717 $\pm$ 0.021
                   & 0.458 $\pm$ 0.105
                   & 0.616 $\pm$ 0.015
                   & 0.577 $\pm$ 0.017          \\
                  MTGB
                   & 0.753 $\pm$ 0.005
                   & 0.748 $\pm$ 0.007
                   & 0.461 $\pm$ 0.101
                   & 0.602 $\pm$ 0.014
                   & 0.573 $\pm$ 0.017          \\
                  R-MTGB
                   & 0.758 $\pm$ 0.006
                   & 0.757 $\pm$ 0.005
                   & 0.459 $\pm$ 0.097
                   & 0.620 $\pm$ 0.013
                   & 0.571 $\pm$ 0.017          \\
                  ST-GB
                   & 0.737 $\pm$ 0.006
                   & 0.726 $\pm$ 0.008
                   & 0.464 $\pm$ 0.108
                   & 0.571 $\pm$ 0.008
                   & 0.577 $\pm$ 0.016          \\
                  ST-LGBM
                   & 0.773 $\pm$ 0.005
                   & 0.767 $\pm$ 0.005
                   & 0.603 $\pm$ 0.125
                   & 0.654 $\pm$ 0.008
                   & 0.530 $\pm$ 0.013          \\
                  TaF-GB
                   & 0.717 $\pm$ 0.011
                   & 0.717 $\pm$ 0.011
                   & 0.147 $\pm$ 0.006
                   & 0.532 $\pm$ 0.007
                   & 0.484 $\pm$ 0.001          \\
                  TaF-LGBM
                   & 0.769 $\pm$ 0.005
                   & 0.769 $\pm$ 0.005
                   & 0.361 $\pm$ 0.148
                   & 0.646 $\pm$ 0.008
                   & 0.488 $\pm$ 0.006          \\
                  \bottomrule
            \end{tabular}
      }
\end{table}

\end{document}